\documentclass[letterpaper, 10 pt, conference]{ieeeconf}  

\IEEEoverridecommandlockouts                              

\usepackage{microtype}
\usepackage{graphicx}
\usepackage{subfigure} 
\usepackage{booktabs} 
\usepackage{diagbox}
\usepackage{hyperref}
\usepackage{pifont}
\usepackage{amsmath}
\usepackage{amssymb}
\usepackage{mathtools}
\usepackage{floatrow}
\usepackage{caption}
\usepackage{bbold}
\usepackage[para]{footmisc}  
\usepackage[ruled, vlined]{algorithm2e}
\usepackage{tabularx}
\usepackage{multirow}
\usepackage{bbm}
\usepackage{wrapfig}

\usepackage{xcolor}
\definecolor{LightCyan}{rgb}{0.88,1,1}
\usepackage{color, colortbl}
\usepackage[subtle]{savetrees}
\newcommand{\squeezeup}{\vspace{-0.6mm}}
\usepackage[capitalize,noabbrev]{cleveref}
\title{\LARGE \bf
Learning a Universal Human Prior for Dexterous Manipulation from Human Preference
}

\author{Zihan Ding$^{1}$, Yuanpei Chen$^{2}$, Allen Z. Ren$^{1}$, Shixiang Shane Gu$^{3}$, Qianxu Wang$^{4}$, Hao Dong$^{4}$ and Chi Jin$^{1}$
\thanks{$^{1}$Princeton University
        }
\thanks{$^{2}$South China University of Technology}
\thanks{$^{3}$OpenAI}
\thanks{$^{4}$Peking University. Correspondence to: Zihan Ding
{\tt\small zihand@princeton.edu}, Chi Jin {\tt\small chij@princeton.edu}.
}
}

\begin{document}

\maketitle
\thispagestyle{empty}
\pagestyle{empty}

\begin{abstract}

Generating human-like behavior on robots is a great challenge especially in dexterous manipulation tasks with robotic hands. Scripting policies from scratch is intractable due to the high-dimensional control space, and training policies with reinforcement learning (RL)  and manual reward engineering can also be hard and lead to unnatural motions. Leveraging the recent progress on RL from Human Feedback, we propose a framework that learns a universal human prior using direct human preference feedback over videos, for efficiently tuning the RL policies on 20 dual-hand robot manipulation tasks in simulation, without a single human demonstration. A task-agnostic reward model is trained through iteratively generating diverse polices and collecting human preference over the trajectories; it is then applied for regularizing the behavior of polices in the fine-tuning stage. 
Our method empirically demonstrates more human-like behaviors on robot hands in diverse tasks including even unseen tasks, indicating its generalization capability.

\end{abstract}

\section{Introduction}
\begin{figure}[htbp]
 \begin{center}
 \includegraphics[width=6cm]{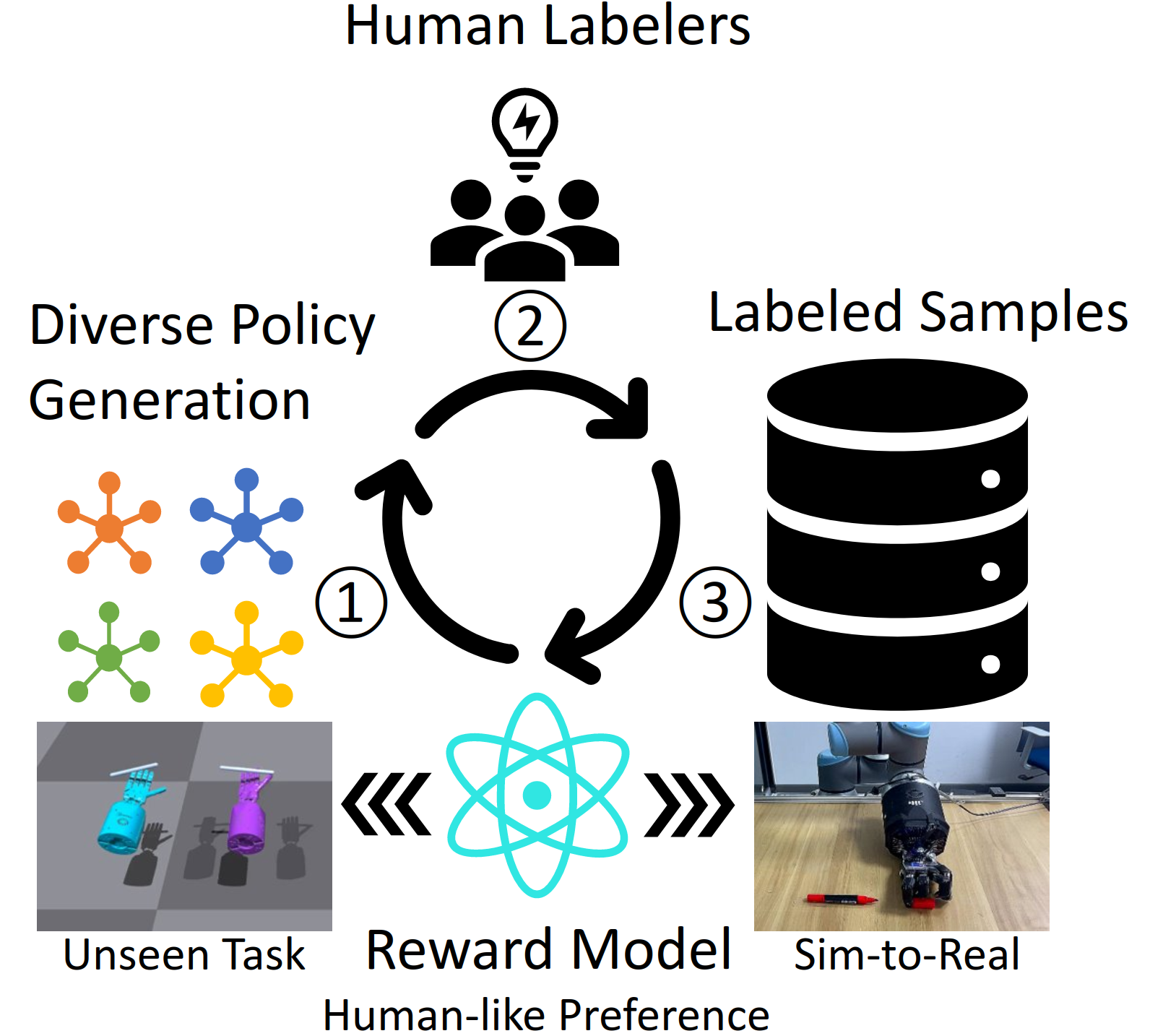}
 \end{center}
	\caption{The proposed method involves an iterative policy fine-tuning procedure with three steps: Step \raisebox{.5pt}{\textcircled{\raisebox{-.9pt} {1}}} is to generate diverse policies across 20 dexterous hand manipulation tasks. Step {\textcircled{\raisebox{-.9pt} {2}}} is to let human labelers provide the preference over trajectories collected from the generated policies. Step {\textcircled{\raisebox{-.9pt} {3}}} is to train the task-agnostic reward model (RM) for human-like behavior using the labeled samples. The polices are fine-tuned in Step \raisebox{.5pt}{\textcircled{\raisebox{-.9pt} {1}}} of the next iteration with the RM. 
 The trained RM can be applied for fine-tuning policies in unseen tasks and sim-to-real transfer.
 }
	\label{fig:overview}
\end{figure}
Dexterous manipulation with multi-finger hands has been gaining popularity in the research community as it enables performing tasks that require dexterity such as rotating objects in-hand or opening a water bottle cap~\cite{akkaya2019solving, chen2022towards}, which are impossible or very difficult for traditional parallel-jaw gripper~\cite{guo2017design}. Model-based motion planning methods are challenging to apply to multi-finger hands due to the high dimension of its action space  (\emph{e.g.}, the hand we use in this work has 30 degrees of freedom) and exponential growth of possible contact modes. Consequently, researchers have been resorting to model-free methods including deep reinforcement learning (RL) with carefully designed reward functions and curriculum for training dexterous manipulation policies \cite{rajeswaran2017learning, chen2022system}.

However, these RL-trained policies tend to generate unnatural and jarring motion. Due to the large action space, the training agent can easily find feasible hand and finger trajectories that satisfy the task completion requirement but does not align with humans' behavioral norms. For example, the robot fingers may twist around each other after throwing an object, or grasping an object in an unnatural way. If we were to deploy these policies in real life, humans might feel uncomfortable and unsafe next to the robots. Humans are also less likely to trust them and question robots' capabilities in solving the tasks. More importantly, human-like behaviors are solutions with higher energy efficiency, better joint protection and limitation satisfaction, indicating the optimality of the control policy. The human-like behaviors are likely to be the movement with minimal energy consumption and motion amplitude. However, this regularization can be hard to manually specify in practice. Therefore, it is challenging to design a method for training multi-finger hand policies to exhibit human-like behavior when performing different tasks. How could we help the robot escape the Uncanny Valley \cite{mori2012uncanny}?

We are inspired by the recent progress in RL with Human Feedback (RLHF), where a reward model (RM) is learned to encode human preferences over data like text generated by large language models (LLMs) \cite{ouyang2022training}. The model is then used as the reward function for RL to fine-tune the original policy. This process helps align the policy with human intent. In this work, we apply the similar idea to regularize the behaviors of policies for dexterous manipulation tasks. With an iterative process of trajectory generation with existing policies, human labeling preferences over robot videos, learning the reward model, and fine-tuning the policies, we gradually improve the human likeness of policies and also the performance across tasks. Compared to using explicit human demonstrations (\emph{e.g.}, teleoperation), which require dedicated equipment (\emph{e.g.}, gloves or other hand tracking devices) and extensive human labor, our approach alleviates the burdens and improves the scalability of encoding human priors in dexterous manipulation training.

Our contributions include (1) proposing a novel pipeline that utilizes human feedback for training diverse multi-finger hand policies and generating human-like behavior in dexterous manipulation tasks; (2) building a platform\footnote{\href{https://sites.google.com/view/openbidexhand}{https://sites.google.com/view/openbidexhand}} for collecting human feedback; (3) training a single task-agnostic reward model for the Shadow Hand robot across 20 dexterous tasks in simulated environment, which demonstrates a $22.3\%$ improvement of preference probability over original RL policies after fine-tuning for four iterations, with evaluations on unseen tasks and real robots.



\section{Related Work}
\subsection{Reinforcement Learning from Human Feedback}
RLHF \cite{akrour2011preference, akrour2012april, griffith2013policy, christiano2017deep, jaques2019way} has been investigated for at least a decade. It is a sub-category of a broader concept called human-in-the-loop learning process \cite{wu2022survey, hejna2022few}. Human feedback data can be essential for some tasks where reward engineering is hard or expensive for RL. 
Research work has been conducted on leveraging human annotated data or demonstrations for robotic control \cite{finn2016guided, cabi2019framework, biyik2022learning}, solving games \cite{ibarz2018reward, vinyals2019grandmaster}, and tuning LLMs~\cite{ziegler2019fine, jaques2019way, stiennon2020learning, madaan2022memory, ouyang2022training}. However, in practice, \textit{human annotation} or \textit{demonstration} can be expensive to acquire. \textit{Human preference} \cite{akrour2011preference, akrour2012april, sadigh2017active, christiano2017deep}, in contrast, is easier to collect as feedback and commonly used in the fields like natural language processing~\cite{ziegler2019fine, ouyang2022training} and robotics~\cite{ibarz2021train}. 


For robotics, human feedback is an important source of information to facilitate the robot learning process \cite{ibarz2021train}. Preference-based learning \cite{sadigh2017active} has been used to provide the reward function for RL agents, with the benefits of better scalability compared to demonstrations. \cite{abramson2022improving} uses the RLHF framework to instruct the learning agents for manipulating objects in a 3D simulated world, and shows improved task success rates over the behavior cloning baseline. 
Few-shot preference learning \cite{hejna2022few} is also investigated with multi-task learning for quick adaptation to new tasks. Moreover, people have explored the combination of demonstrations and preferences as guiding signals for robots \cite{biyik2022learning}. As discussed in Appendix~\ref{app:approach_hf}, our applied RLHF approach distinguishes with other previous work on robot hand manipulation with human feedback for the ease of data collection and less engineering effort. With the RLHF approach, the data collection and feedback time are significantly reduced without expensive human demonstrations. Preference over videos requires small amount of efforts from humans thus can produce a large amount of labeled data.

\subsection{Natural Human Behavior in Robotic Manipulation}
There is a branch of work leveraging human demonstrations and imitation learning \cite{rajeswaran2017learning, christen2019guided, sivakumar2022robotic, mandikal2022dexvip, jiang2019synthesis, ren2021generalization, alakuijala2021residual, arunachalam2022dexterous, du2022multi, lopez2023dexterous, dasari2022learning} for robotics. 
The grasping operation is one key step for dexterous manipulation with hands. For human-like grasping \cite{mandikal2022dexvip, du2022multi, zhu2021toward, ye2022learning, wang2022dexgraspnet, sievers2022learning, qi2023hand}, previous works have used carefully engineered loss function~\cite{zhu2021toward, sievers2022learning, qi2023hand}, or optimization under reachability and collision constraints~\cite{wu2022learning}, or leveraging heavy human demonstrations~\cite{ye2022learning, wang2022dexgraspnet, du2022multi} with DexYCB dataset~\cite{chao2021dexycb}, DexGraspNet~\cite{wang2022dexgraspnet} or DEXVIP~\cite{mandikal2022dexvip}. Unidexgrasp~\cite{xu2023unidexgrasp} demonstrates universal grasping across objects with dexterous hands but without further task solving, while Bi-DexHands~\cite{chen2022towards} provides a simulated task solving platform without considering the human-likeness in the designed reward functions. Our work distinguishes with the above work in several aspects: (1) no demonstration data is used in our method, but only human preferences over videos are collected; (2) our work is not only useful for grasping, but also for a broader category of dexterous manipulation tasks including turning water bottle caps and opening doors; (3) our work focuses on improving the human-likeness of the behaviors instead of just the task completion rate.

 \begin{figure*}[htbp]
	\centering\includegraphics[width=0.8\textwidth]{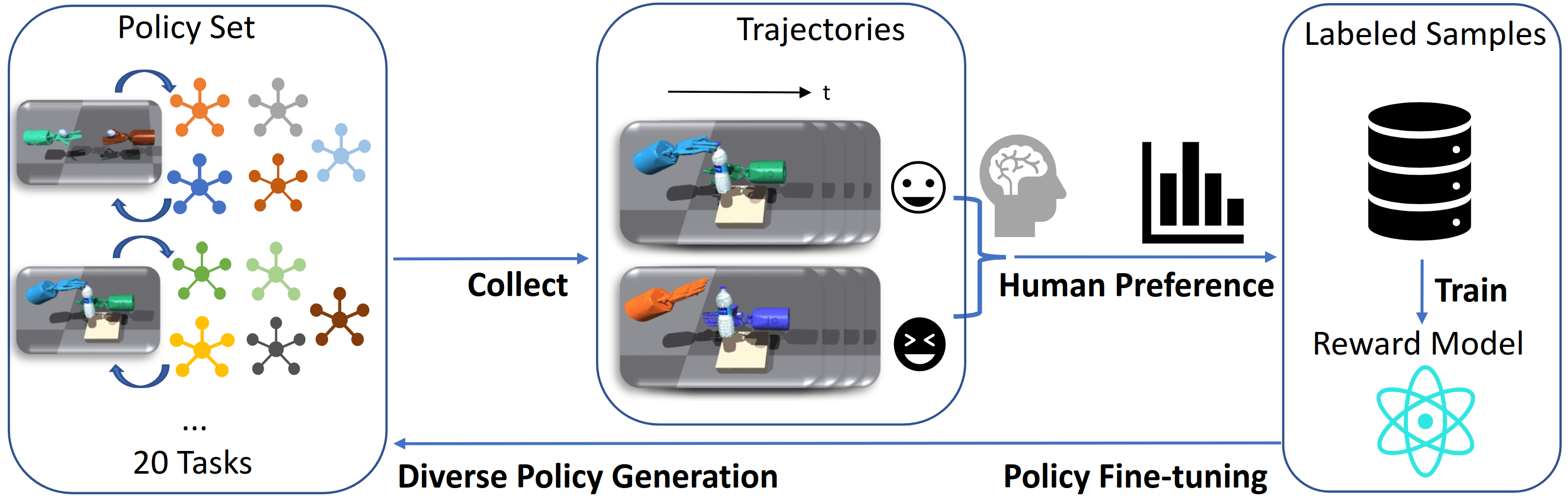}
	\caption{An overview of the proposed framework includes five steps in iteration: 1. Diverse policy training; 2. Trajectories collection from diverse policy set; 3. Human preference over collected trajectory pairs; 4. Reward model training with the labeled sample from human preference; 5. Policy fine-tuning for re-generating diverse polices using reward model.}
	\label{fig:detailed}
\end{figure*}

\section{Methodology}
 
Although human data is already widely adopted in tuning the performance of the learning agents in previous work, we highlight the rationale of our design choices given key properties of the dexterous robot manipulation problem:
(1). Using human preferences: instead of using demonstrations, which can be expensive and requiring human expertise, the human preferences over video are much cheaper to get. 
(2). Using a single task-agnostic reward model: human-like behavior regularization should be task-agnostic, as an approximation of a combination of principles involving energy minimization, lowest frictions, and avoiding violating the joint limits, which can be hard to specify in task rewards. Fig.~\ref{fig:detailed} shows the proposed framework as an iterative policy fine-tuning process from human-preference.


\subsection{Preliminary}
\label{sec:prim}
We use Proximal Policy Optimization (PPO) \cite{schulman2017proximal} algorithm in our experiments. The parameterized policy $\pi_\theta$ is optimized with the loss: $\mathcal{J}(\pi_\theta; r)=-\mathbb{E}_{s\sim \rho_\pi, a\sim\pi}[\min(R_\theta A(s,a), \text{clip}(R_\theta, 1-\epsilon, 1+\epsilon)A(s,a))]$,  where ratio $R_\theta=\frac{\pi_\theta(a|s)}{\pi_{\theta_\text{old}}(a|s)}$ and $\pi_{\theta_\text{old}}$ is the previous model for generating training samples, the advantage function $A^{\pi_\theta}(s,a)$ is estimated as $\sum_t \gamma^t r_t(s,a)-V^{\pi_\theta}(s)$, with state-value function $V^{\pi_\theta}(s)$. $r$ is the reward function and $\rho_\pi$ is the state-visitation distribution by policy $\pi$. 

\subsection{Diverse Policy Generation}
\label{sec:diverse}
To allow collecting human preferences over trajectories, we first need to generate diverse task policies for humans to choose from. We apply the PPO algorithm with additional diversity loss on constraining the action log-probabilities. Without the diversity loss, policies trained even with different random seeds are likely to collapse to very few modes or a single mode in terms of behavior. 

The diversity loss for updating current policy $\pi_\theta$, given an existing policy set $\mathcal{S}$, is:
\begin{equation}
    l_d(\mathcal{S}) = \frac{1}{|\mathcal{S}|}\sum_{i=1}^{|\mathcal{S}|}\mathbb{E}_{a\sim\pi_\theta(\cdot|s)}[\frac{-1}{1+\log \pi_i(a|s)}]
    \label{eq:add_loss}.
\end{equation}

The pseudo-code for generating a diverse policy set $\mathcal{S}$ on one task is shown in Alg.~\ref{alg:diverse_ppo}. For generating diverse polices before the first iteration of fine-tuning, the reward function in objective $\mathcal{J}(\pi;r)$ is just the task reward: $r=r_\text{Task}$. The effects of diverse policy generation are evaluated in Sec.~\ref{sec:exp_details}.

\begin{algorithm}[htbp]
\small
\caption{\textcolor{cyan}{Diverse} Policy Generation with PPO}
Initialize policy set: $\mathcal{S}=\emptyset$.\\
\For{training iteration $i = 1,\ldots,N$}{
 Initialize policy $\pi_i$, value $V_i$ in PPO.\\
\While{$\pi_i$ \text{ not converged }}{
Run policy $\pi_i$ to collect samples $\{(s,a,r,g,s^\prime)\}$.
Update policy $\pi_i$ with loss: $-\mathcal{J}(\pi_i)+\xi \color{cyan}{l_d}$. (Eq.\eqref{eq:add_loss})
Update value $V_i$ as standard PPO.
}
Update policy set: $\mathcal{S}=\mathcal{S}\bigcup \{\pi_i\}$.
}
\vskip -0.05in
\label{alg:diverse_ppo}
\end{algorithm}

\subsection{Human Preference Collection}
\label{sec:human_preference}

Our proposed the RLHF approach is compared with other commonly seen approaches for leveraging human feedback in robot manipulation tasks in Appendix~\ref{app:approach_hf} Table~\ref{tab:compare_feedback}. These approaches include learning from human demonstrations in real world ~\cite{christen2019guided} and imitating behaviors from human with online videos \cite{sivakumar2022robotic}. 

The desiderata of an approach for leveraging human feedback involves time efficiency for getting both data and human feedback, the human resources and special requirement (\emph{e.g.}, whether an expert is required), the engineering difficulty of implement the methods. Due to embodiment mismatch of human bodies and the robots, a re-targeting procedure is usually required for imitating the behaviors from human directly, either via live human demonstrations or videos from a human. This can not only increase the engineering difficulty for using human feedback data, but also induce errors and uncertainties in the pipeline. Our proposed framework with human feedback over trajectory videos has the benefits of requiring a medium amount of time for video data collection and a relatively small amount for human feedback collection. Since all trajectories are collected in simulation with policies trained using RL algorithms, the amount of available data is large. Moreover, any normal person can serve as a labeler after very brief instruction and familiarization of the user interface, while collecting human demonstrations usually requires a human expert wearing certain equipment~\cite{christen2019guided}.

\subsection{Reward Model Training}
\label{sec:rm_train}
The reward model $r_\text{HF}(\cdot; \phi)$ is formulated as the \textit{Bardley-Terry-Luce}~\cite{bradley1952rank} model (detailed descriptions in Appendix~\ref{app:rm_form}) for pairwise-sample comparison, inspired from \cite{abramson2022improving}. However, we directly compare two trajectories generated from different policies, and train the RM with the loss:
\begin{equation}
    l_\text{RM} = \mathbb{E}_{\tau_1, \tau_2 \sim \mathcal{D}}\big[-\log \sigma\big(r_\text{HF}(\tau_1; \phi) - r_\text{HF}(\tau_2; \phi)\big)\big], \\
    \tau_1 \succ \tau_2,
\label{eq:rm_loss}
\end{equation}
where $r_\text{HF}(\tau_i; \phi)$ is the reward model , $\tau_i=[s_1, a_1, \dots, s_M, a_M], i\in\{1,2\}$ is the stacked state-action pairs of length $M$, `$\succ$' is the preference relationship. The training dataset $\mathcal{D}=\bigcup_{i\in\mathcal{S}, j\in \mathcal{T} } \mathcal{D}_{\pi_{i,j}}$ contains the human preferences collected over sample trajectories, which are generated using the current set of diverse policies $\mathcal{S}$ up to the current training iteration, across all tasks in set $\mathcal{T}$. Each trajectory of length $H$ is transformed into $(H-M+1)$ consecutive samples of length $M$ through a sliding window.

\subsection{Fine-tuning Task Policy with RM}
\label{sec:policy_rm}
The objective for fine-tuning the task policy with learned reward model $r_\text{HF}$ is $\mathcal{J}(\pi_\theta; \tilde{r})$ with:
\begin{equation}
    \tilde{r} = r_\text{Task}(s_t) + \alpha\cdot c \cdot r_\text{HF}(\tau_t),
    \label{eq:rm_obj1}
\end{equation}
where $c=|\overline{r_\text{Task}}|$ is a scaling term tracking the magnitude of averaging task reward over time.
$\tau_t$ is the stacked state-action pairs at time-step $t$. The score from the reward model $r_\text{HF}$ serves as an additional regularization term for the task reward, with a proper scaling. In our experiments, this objective is shown to be effective for tuning the policy behaviors to follow human preference.


\section{Experiments}
\subsection{Task and Environment Settings}
\paragraph{Bi-DexHands Environments.}
Bi-DexHands~\cite{chen2022towards} is a collection of bimanual dexterous manipulation tasks and reinforcement learning algorithms, aiming at achieving human-level sophistication of hand dexterity and bimanual coordination. 20 tasks are used in our experiments. Most tasks involve two Shadow hands and different manipulated objects. Each hand has 24 degrees of freedom (DoF), which leads to high-dimensional observation and action spaces. More details about observation space, action space, and reward design for each task in Bi-DexHands are referred to Appendix \ref{app:tasks}.

\subsection{Experimental Details}
\label{sec:exp_details}
\paragraph{Policy Generation and Data Collection.}
For each iteration, we train 10 policies for each task with different random seeds, with the diverse policy generation loss and the learned RM from last iteration (except for the first iteration). For the first iteration, each policy is trained for 20000 episodes to achieve the task completion, and in subsequent iterations, the policies are initialized with checkpoints from the previous iteration and fine-tuned for 5000 episodes. To ensure the policies can reasonably complete the tasks and thus then used used for trajectory collection, in the first iteration, we visualize the task performance for polices at all checkpoints, spread by a 1000-episodes interval, and choose only those checkpoints with successful task completion. In the trajectory collection phase, we collect 5 trajectories with each policy checkpoint in simulation. About 12300 trajectories (tasks$\times$seeds$\times$checkpoints$\times$rollouts) across 20 tasks are collected for the first iteration. After reviewing the videos, we decide to discard three tasks \texttt{SwingCup}, \texttt{Kettle} and \texttt{DoorCloseOutward} due to the difficulty in task completion. For rest of the iterations, about 4100 trajectories\footnote{Fewer checkpoints are saved compared to the first iteration.} are generated in each iteration. The hyperparameters for policy generation with PPO algorithm are shown in Appendix~\ref{app:hyper}.

We provide visualization of t-SNE~\cite{van2008visualizing} plots for trajectories of four tasks, as in Appendix~\ref{app:diversity}. 
From the results we can see the different policies well separated in the state space. This benefits the downstream procedure for human preference data collection. In Tab.~\ref{tab:diversity_compare}, we compare the proposed approach with policy entropy method and without any bonus for diversity to justify our design choice with higher success rate and larger diversity. More details refer to Appendix~\ref{app:diversity}.

\begin{table}[t]
\centering
\caption{Diversity and success rate of three methods for policy training on three tasks.}
\label{tab:diversity_compare}
\resizebox{\columnwidth}{!}{ 
\begin{tabular}{c|ccc|ccc}
\toprule
       &  \multicolumn{3}{c|}{Success Rate} & \multicolumn{3}{c}{Diversity} \\ \hline
      \backslashbox{Task}{Method} & No Bonus & Entropy & Ours & No Bonus & Entropy & Ours \\ \hline
     \texttt{HandOver}  &  $\mathbf{85.4\pm 26.5}$   & $77.3\pm 23.3$     &  $85.2\pm 26.3$ & $53.0$ & $55.0$ & $\mathbf{58.7}$ \\ \hline
     \texttt{Pen} & $74.8\pm 38.9$  & $70.4\pm 36.7$      &  $\mathbf{89.3\pm 27.7}$ & $77.9$  & $75.9$ & $\mathbf{78.4}$  \\ \hline
     \texttt{CatchOver2}  &  $76.2\pm 23.5$ & $65.4\pm 26.7$      &  $\mathbf{81.3\pm 24.5}$  & $53.8$ & $54.7$ & $\mathbf{57.3}$   \\ \midrule
\end{tabular}
}
\end{table}

\paragraph{Human Feedback Collection.}
\begin{figure}[htbp]
     \includegraphics[width=\columnwidth]{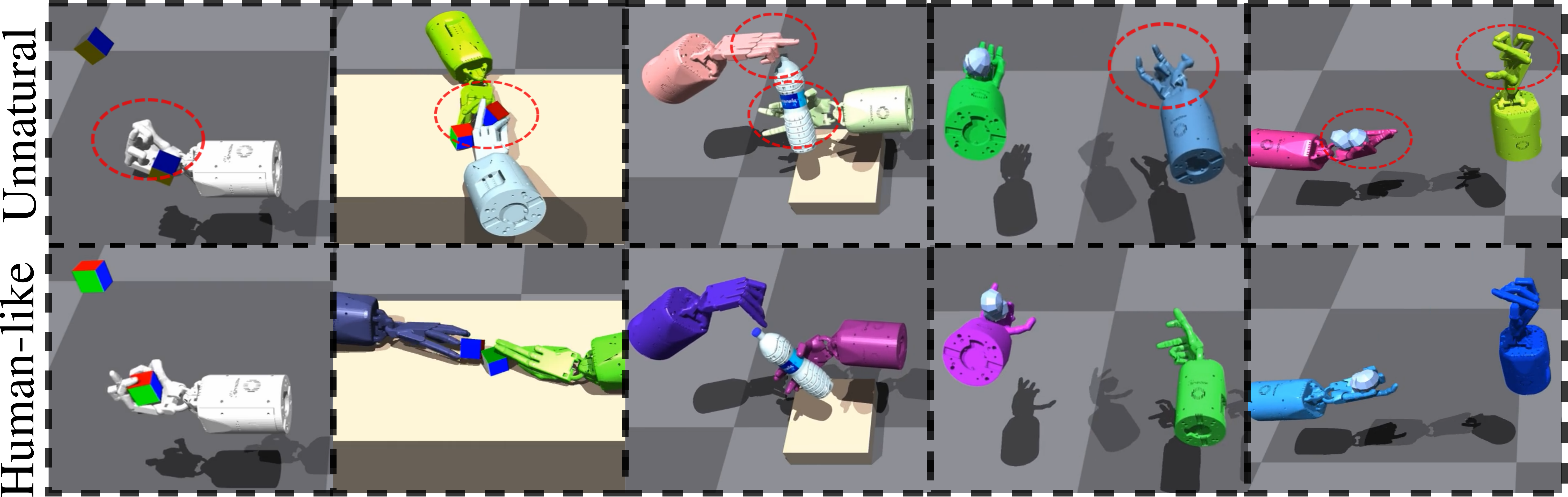}
     \caption{Comparison of unnatural (top) behaviors with original RL polices and human-like (bottom) behaviors after tuning with RM for four iterations on five tasks (left to right): \texttt{ShadowHand}, \texttt{PushBlock}, \texttt{BottleCap}, \texttt{CatchUnderarm} and \texttt{CatchOver2Underarm}. Red circles indicate the unnatural poses of the hand, including twisted fingers, over-stretched hand poses, weird correlated positions of joints.}
	\label{fig:vis_behav}
\end{figure}

We recruit five human labelers providing preferences with the feedback collection interface we build. A total of 1000 feedback over trajectory pairs are collected for each iteration, which is then converted into $1\times10^5\sim2\times10^5$ labeled samples (This number varies due to different trajectory lengths). Each feedback takes about 10-20 seconds and the data collection can be finished in several hours. Each preference is the choice over `Left', `Right' and `Not Sure' based on the given two side-by-side trajectory videos for the same task. The `Left' indicates that the trajectory on the left shows a more human-like behavior than the right and vice versa. The preference data is processed to be stacked state-action pairs with a sliding window on each trajectory. The window size is chosen to be 8 in our experiments, which corresponds to about 0.33 seconds in videos. The processed data is used for training the RM with loss as Eq.~\eqref{eq:rm_loss}. Fig.~\ref{fig:vis_behav} visualizes a side-by-side comparison of frames in five tasks\footnote{\texttt{ShadowHand}, \texttt{BlockStack}, \texttt{BottleCap}, \texttt{CatchAbreast}, \texttt{CatchOver2Underarm}}, showing the differences of unnatural and human-like behavior.

\begin{figure}[htbp]
	\centering
     \includegraphics[width=\columnwidth]{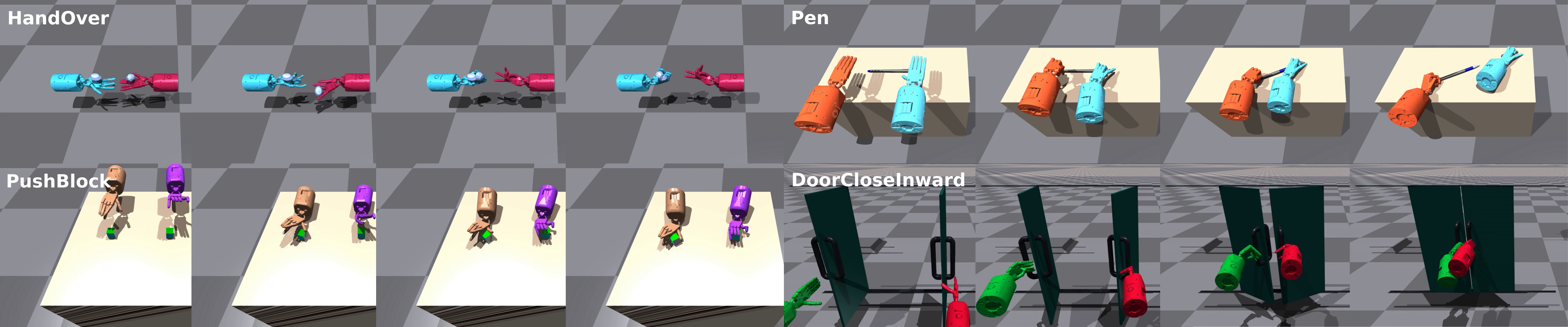}
	\caption{Visualization of hand and object trajectories in four tasks. From left to right it shows the procedure of the task completion.}
	\label{fig:vis_traj}
\end{figure}

\paragraph{Feedback in the Format of Preferences over Trajectories.}
Previous works have used human preference over video clips~\cite{christiano2017deep, abramson2022improving} or the whole trajectories~\cite{akrour2011preference, akrour2012april}. Compared with whole trajectories, video clips are shorter and therefore more time efficient for label collection. However, the previous work with video clips~\cite{christiano2017deep, abramson2022improving} focus on the improvement on the task completion for Atari games, MuJoCo or Playhouse environments, which are all long-horizon tasks. Although solving tasks in the Bi-DexHands environment require complex dexterous manipulation, the intentions of the robots are usually straightforward and the tasks have relatively short horizons, ranging from 20 to 600 timesteps (details in Appendix \ref{app:traj_length}). Clipping the trajectories in this environment can increase difficulties for the labelers to provide usefulpreferences. Fig.~\ref{fig:vis_traj} visualizes the trajectories for for tasks: \texttt{Pen}, \texttt{HandOver}, \texttt{PushBlock} and \texttt{DoorCloseInward}.

\vspace{-5.mm}

\paragraph{Models and RM Training.}
The RM in our experiments is parameterized by a fully-connected neural networks with 512-512-512-128-32 hidden units and Tanh activation function for both hidden layers and the output layer.  The input shape of RM is $(d_\mathcal{O}+d_\mathcal{A})\times M$, where $d_\mathcal{O}=24$ is the dimension of joint positions of the full robot hand\footnote{Only $\mathcal{J}_p$ in $\mathcal{O}_\text{left hand}$ and $\mathcal{O}_\text{right hand}$, see Appendix~\ref{app:obs_space}.} and $d_\mathcal{A}=20$ is the dimension of the proactively driven joint actions\footnote{Only $\mathcal{A}_\text{left hand}$ and $\mathcal{A}_\text{right hand}$, see Appendix~\ref{app:act_space}.} and $M=8$ is the number of stacked frames. The optimization process uses Adam optimizer \cite{kingma2014adam} for minibatch stochastic gradient descent of 50000 epochs, with batch size 4096, learning rate $1\times10^{-3}$ and the multiplicative learning-rate scheduler StepLR in PyTorch with step size as 1000 and gamma as 0.5. For each iteration, the newly collected data is appended to the previous data as a whole for training a new RM initialized with the checkpoint of the last iteration.

We want to emphasize our choices of the input for the RM as a universal module for regularizing the behavior of Shadow Hand. As opposed to use the entire observation and action for the dual-hand tasks (detailed in Appendix~\ref{app:tasks}, which can be of hundreds of dimensions, we choose only the critical joint states and hand-only actions as the inputs to the RM. This choice of design benefits from several perspectives: (1) reduction of input dimensions increases the training and inference efficiency, without the need of using a much larger neural network; (2) this also helps to reduce the required number of samples for training the RM; (3) since the observations and actions only involve the joint states on Shadow Hands, it is \textit{task-agnostic}. The human-like behavior is assumed to be mostly affected by the relative motions of fingers on hand instead of the overall movement of the hand basis for these dexterous manipulation tasks.

\subsection{Reward Model Evaluation}


After collecting and summarizing the human feedback on sampled trajectories, we introduce the human preference score to quantify the preference over policies.

\textbf{Human Preference Score} is the metric showing the preference of humans over the policy sets.
\begin{equation}
   c_\text{HF}= \frac{1}{M}\sum_{k=1}^M \big[ \mathbb{1}(\tau_{i,k}\succ \tau_{j,k}) - \mathbb{1}(\tau_{j,k}\succ \tau_{i,k})\big], \tau_{i,k}\in\mathcal{T}_i, \tau_{j,k}\in\mathcal{T}_j \notag,
   \label{eq:human_prefer_score}
\end{equation}
where $\mathbb{1}(\tau_{i,k}\succ \tau_{j,k})$ indicating the labeler's preference of trajectories $\tau_{i,k}$ over $\tau_{j,k}$ and vice versa. $\mathcal{T}_i$ is the set of trajectories collected with the policy set indexed by $i$. $\tau_{i,k}$ and $\tau_{j,k}$ are randomly selected from the corresponding sets. $M$ is the set of paired samples for labelers to provide preference. This score rules out the samples labeled as `Not Sure' in the data. The results showing the consistency of RM and human preference scores are shown in Fig.~\ref{fig:rm_compare}, with more details in Appendix~\ref{app:rm_eval}.

\begin{figure}[htbp]
\centering\includegraphics[width=0.9\textwidth]{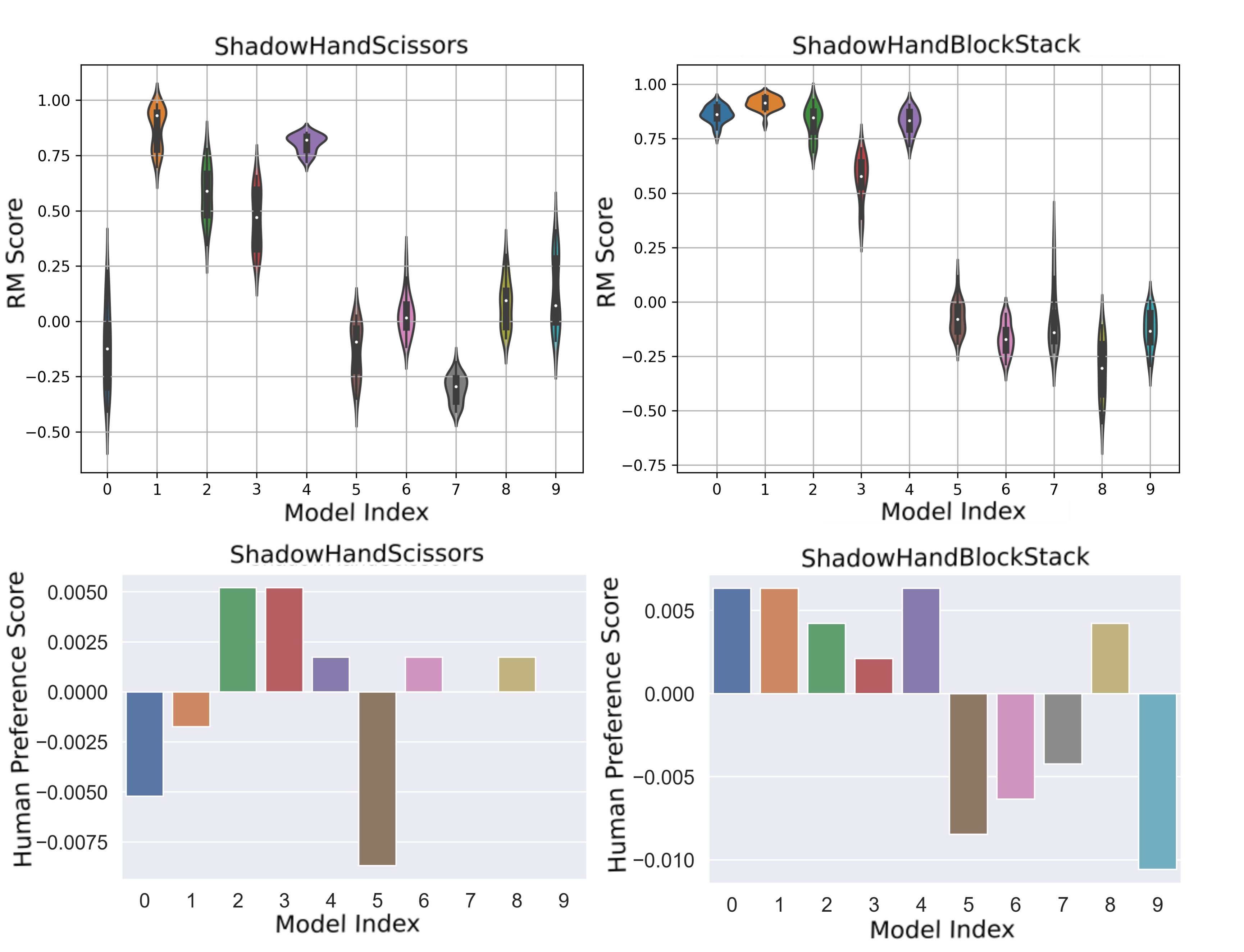}
	\caption{Comparison of the accumulative rewards over trajectories by RM (top) and human preference scores (bottom). }
	\label{fig:rm_compare}
\vskip -0.1in
\end{figure}


\subsection{Human-like Robot Polices with RLHF}
\paragraph{Preference Results.}

\begin{table}[htbp]
\caption{Preference results.}
\vskip -0.05in
\label{tab:rm_results}
\centering
\footnotesize
\resizebox{\columnwidth}{!}{ 
\begin{tabular}{p{2cm}ccc}
\toprule
        & \multirow{1}{*}{\textbf{Policy+RM}} & \multicolumn{1}{c}{Original Policy} & \multicolumn{1}{c}{Not Sure} \\ \hline
      \textbf{First} Iteration (Seen Tasks) &  $25.1\%$           & $22.2\%$   & $52.7\%$ \\ \hline
      \textbf{Final} Iteration (Seen Tasks) &  $35.7\%$            & $13.4\%$   & $50.9\%$ \\ \hline
      \textbf{Final} Iteration (Unseen Tasks) &  $24.9\%$            & $13.0\%$   & $62.1\%$ \\ \midrule
\end{tabular}
}
\end{table}
\squeezeup
After evaluating the learned RM, we apply the task-agnostic RM as an additional term in the task reward to fine-tune the task-specific policies, as introduced in Sec.~\ref{sec:policy_rm}. The RM training and policy fine-tuning process are iterated for four times in our experiments. For the first and final iterations (the fourth), Table~\ref{tab:rm_results} shows the results of preference evaluation for polices with RM fine-tuning and without it. The numbers in the table indicate the percentage of evaluation trials for each case. For example, in the row `First Iteration', $25.1\%$ of evaluation trials show a preference of `Policy+RM' over `Original Policy', and $22.2\%$ the opposite. With $52.7\%$ probability, the human labelers are not sure on the preference. `First Iteration' indicates the comparison after one iteration of RM training. `Final Iteration' indicates the comparison after multiple iterations of RM training and policy tuning. As shown in the table, the effects of RM fine-tuning can be insignificant with only one iteration. However, the preference over polices fine-tuned with the RMs increases from $25.1\%$ to $35.7\%$ with more iterations of RM training and policy fine-tuning. In Appendix~\ref{app:preference_breakdown}, the breakdown results for the preference over each task are shown. 

\begin{table}[htbp]
\caption{Average success rates over seen tasks (17) and unseen tasks (4). Breakdown results in Appendix~\ref{app:success}.}
\label{tab:success}
\scriptsize
\resizebox{\columnwidth}{!}{ 
\scriptsize
\begin{tabular}{c|cc}
\toprule
     Average Success Rate &  \multirow{1}{*}{\textbf{Policy+RM}} & \multicolumn{1}{c}{{Original Policy}}  \\ \hline
      Seen Task  &  $0.47$        & $0.55$   \\ \hline
     Unseen Task  &  $0.47$        & $0.48$   \\ \bottomrule
\end{tabular}
}
\end{table}


The high probability of uncertainty (`Not Sure') is within our expectation, since the comparison of human-like behaviors can be very subtle in some cases and we ask the labelers to only label a clear preference when they are certain to see a significant difference between the compared trajectories.
In the results, the policies are evaluated with human labelers using the same interface as the one for providing preference feedback. The same people (five labelers) for providing feedback are providing evaluations. 500 trials of evaluation are provided for each comparison. The time for providing evaluation is the same as for providing feedback, as about 10-20 seconds per evaluation.

\paragraph{Success Rates.}
\begin{figure*}[htbp]
\floatbox[{\capbeside\thisfloatsetup{capbesideposition={left,top},capbesidewidth=0.3\columnwidth}}]{figure}[\FBwidth]
{\caption{Visualization of trajectories with action sequences using fine-tuned polices with the RM for two tasks \texttt{Pen} (top) and \texttt{Relocate} (bottom) in simulation-to-reality experiments.}}
	{\includegraphics[width=12cm]{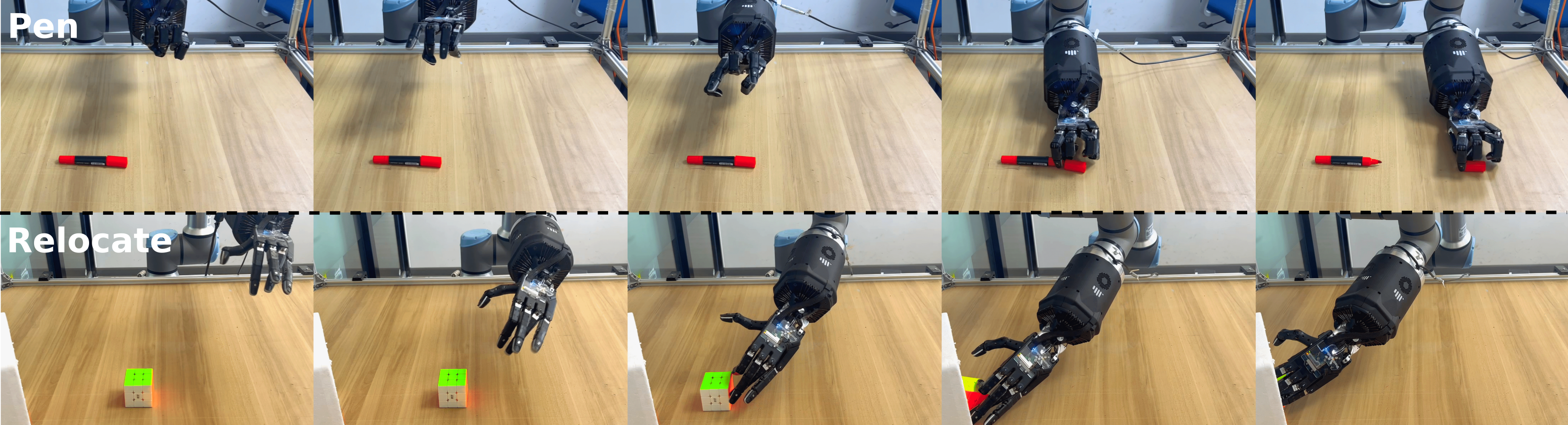}
 \label{fig:real_exp}}
\end{figure*}
Tab.~\ref{tab:success} summarizes the success rates for all 17 tasks (seen) finally adopted for training RM and tuning polices. The learning curves for all tasks are shown in Appendix~\ref{app:learning_curve}. Each task is evaluated with policies trained with 10 random seeds, and the results in the table show the means and standard deviations of the final success rates over the 10 runs. It shows that success rates are negatively affected by the RM fine-tuning process by $8\%$. This is within our expectation since the RM serves as an additional regularization term for the original task reward. There is the trade-off between the task completion and the human-like behaviors in our settings. Unlike some previous work~\cite{christiano2017deep} using RM for providing task completion guidance, in our setting the RM is used for calibrating the robot behavior. We admit that this may not be always aligned with the task completion objective, especially in some subtle cases requiring human-likeliness as constraints. This issue is further investigated with different choices of the RM objective and the scaling hyperparameter for using the RM in fine-tuning, and the results are discussed in Appendix~\ref{sec:caveats}. It may be noticed that some tasks have relatively low success rates, like \texttt{LiftUnderarm} and \texttt{TwoCatchUnderarm}. These tasks are relatively hard to solve due to the strict success conditions in the environments and no usage of any human demonstration. The \texttt{LiftUnderarm} task requires two hands collaboratively lifting up a heavy pot to a certain height, and the \texttt{TwoCatchUnderarm} task requires two hands each to throw a ball and catch the ball thrown by the other hand. These tasks themselves are very challenging for RL polices.

In Tab.~\ref{tab:compare_re}, we additionally compare our approach with reward engineering (RE)~\cite{sievers2022learning, qi2023hand} approach in terms of task success rates, which demonstrates that naive RE tuning with torque and energy consumption penalties can hurt task completion significantly, not to mentioning the difficulty of tuning coefficients for all penalty terms in such a high-dimensional control space. More details about this comparison are discussed in Appendix~\ref{app:compare_re}. 

\begin{table}[htbp]
\caption{Success rates on three training tasks.}
\centering
\label{tab:compare_re}
\resizebox{\columnwidth}{!}{ 
\begin{tabular}{c|ccc}
\toprule
       Task &  \multirow{1}{*}{\textbf{Policy+RM}} & \multicolumn{1}{c}{Original Policy} & \multirow{1}{*}{Policy+RE}  \\ \hline
     \texttt{CatchOver2Underarm}  &  $0.62\pm0.25$        &  $0.74\pm0.24$   & $0.63\pm0.26$  \\ \hline
     \texttt{HandOver}  &  $0.79\pm 0.31$        &  $0.83\pm0.24$   & $0.78\pm0.30$     \\ \hline
     \texttt{BottleCap}  &  $0.58\pm0.39 $        &  $0.63\pm0.28$    & $0.00\pm0.00$    \\ \hline
     \texttt{Pen}  &  $0.78\pm0.29$        &  $0.66\pm0.43$    & $0.38\pm0.34$   \\ \hline
     Total  &  $0.69$        & $0.71$   & $0.45$    \\ \midrule
\end{tabular}
}
\vskip -0.1in
\end{table}

\subsection{Generalization to Unseen Tasks}

\begin{figure}[htbp]
	\centering
     \includegraphics[width=0.95\columnwidth]{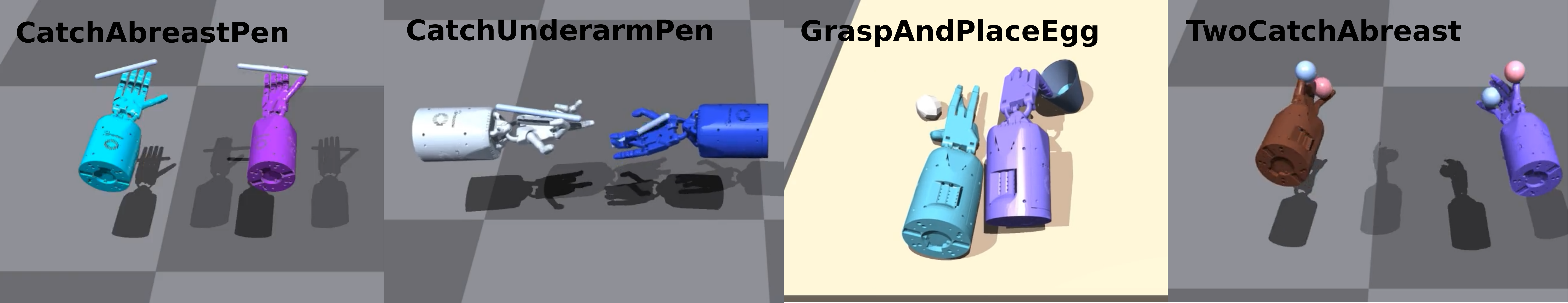}
	\caption{Visualizing four unseen tasks.}
	\label{fig:vis_unseen}
\end{figure} 

With the trained RM in the last iteration, we further test its generalization capability in the four unseen tasks as shown in Fig.~\ref{fig:vis_unseen}, \texttt{CatchUnderarmPen}, \texttt{CatchAbreastPen}, \texttt{TwoCatchAbreast}, \texttt{GraspAndPlaceEgg}, where the manipulated objects or the movement objectives are changed. For each task, original policies are trained under 10 random seeds for 50000 episodes. For fine-tuning, the RM is used to fine-tune ten polices for 20000 episodes in each task. The preference results are shown in Tab.~\ref{tab:rm_results} (with breakdown results in Appendix~\ref{app:preference_breakdown} Tab.~\ref{tab:unseen_prefer_breakdown}). The preference scores of the RM on the unseen tasks are lower than on training tasks, but RM still outperforms original policies by $11.9\%$, which shows the generalized improvement of using RM for fine-tuning polices on unseen tasks. The success rates for unseen tasks are shown in Tab.~\ref{tab:success}. We notice that the success rates even increase in two of the unseen tasks (\texttt{CatchAbreastPen} and \texttt{CatchUnderarmPen}) with RM fine-tuning. For the hard task \texttt{TwoCatchAbreast}, the success rate is zero given the strict success condition on target ball positions, but the two hands also learn reasonable ball throwing behaviors therefore the preferences are provided on these behaviors.


\subsection{Real Robot Experiments}
\label{sec:real}
To demonstrate the benefits of human-like behaviors on simulation-to-reality process, we further test the applied RLHF approach in real world robots with policies fine-tuned in simulation on tasks: \texttt{Pen} and \texttt{Relocate}. The RM trained in previous experiments for four iterations is applied for tuning the new task policies here in simulation. Still, no demonstration is used for the real robots. The real-robot experimental setup includes a Shadow Hand mounted at the end of a UR10e robotic arm with a control frequency of 10Hz. The policies trained with and without RM fine-tuning generate action sequences to be executed on the real robot for accomplishing the tasks. As shown in Fig.~\ref{fig:real_exp}, policies fine-tuned with RM are found to have smooth operation trajectories and benefit the simulation-to-reality transfer. The real-robot experiments are important to show the effectiveness of the human-like behaviors via our method not only in the simulation, but also help with the robust transfer to real world. More details refer to Appendix~\ref{app:real_exp}.


\section{Conclusions and Limitations}
\label{app:limit}
In this work we propose using human preference feedback to learn a universal human prior for multi-finger in-hand manipulation over diverse dexterous tasks. With an iterative process of policy learning, feedback collection, and human reward model learning, the proposed method based on RLHF can significantly improve the human likeness of the hand trajectories without hurting too much of the task performance. Experiments on unseen tasks and real robots demonstrate its generalization capability.

There are limitations and potential extensions for current method and experiments. The subtleties of in-hand manipulation can affect the human likeness thus make it hard for providing human preference over videos. The amount of human preference data is limited in current experiments, and more data is expected to further improve the learned RM for calibrating human-like behaviors. There are some caveats for applying the learned RM in tuning policies as summarized in Appendix~\ref{sec:caveats}. Although human preferences are cheaper than human demonstrations, we believe there are approaches to further alleviate the required human efforts by changing the fashion of feedback or improving the data efficiency by inspecting the preference data and leveraging it with prioritized sampling mechanism. These are promising directions to explore in the future work.

\bibliography{ref}
\bibliographystyle{IEEEtran}

\clearpage
\onecolumn
\section*{APPENDIX}
\section{Approaches for Human Feedback in Dexterous Manipulation}
\label{app:approach_hf}
\begin{table*}[htbp]
\caption{Comparison of different approaches for using human feedback in robot hand manipulation.}
\label{tab:compare_feedback}
\centering
\resizebox{\columnwidth}{!}{ 
\begin{tabular}{p{2.4cm}cccccc}
\toprule
        & \multirow{1}{*}{Data Collection Time} & \multicolumn{1}{c}{Feedback Time} & \multicolumn{1}{c}{Data Amount} & \multicolumn{1}{c}{Expert}  & \multicolumn{1}{c}{Engineering Difficulty} & \multicolumn{1}{c}{Re-targeting}  \\ \hline
      \rowcolor{LightCyan} 
      RLHF (Ours) &   Medium            & $10\sim20 s$ & Large  & No &  Low  & No\\ \hline
      Demo~\cite{christen2019guided}  &  Large  &  Minutes  & Small & Yes & High & Yes\\  \hline
      Video~\cite{sivakumar2022robotic, mandikal2022dexvip}  &  {$\sim 0$} &  {$10\sim20 s$} & {Very Large} & {No} & {High} & {Yes}\\ \bottomrule
\end{tabular}
}
\end{table*}
Table~\ref{tab:compare_feedback} compares different approaches for leveraging human feedback in robot hand manipulation tasks from several aspects. Although not rigorously quantified, the contents in the table show the general requirements for our approach, learning from human demonstrations and learning from videos. The proposed RLHF approach only requires data collection in simulation with trained policies, and each human feedback is the evaluation time of human labeler watching the pairwise videos. Although some approaches may report to use only a few demonstrations~\cite{sivakumar2022robotic}, it can be designed for task-specific purpose, while our approach tries to provide a universal human-like prior across tasks. The expert column indicates whether a person familiar with the experimental equipment (if any) and the procedure is required to provide the feedback. The engineering difficulty includes the amount of engineering work required to prepare for the feedback collection process, as well as additional data processing like the re-targeting process due to the human-robot embodiment mismatch~\cite{christen2019guided}.

\section{Reward Model Formulation}
\label{app:rm_form}
\paragraph{\textit{Plackett-Luce} Model and \textit{Bardley-Terry-Luce} Model}
The \textit{Plackett-Luce} (PL) model is a general preference model over length-$K$ ranking. Specifically, the probability of a permutation (or ranking) $\sigma$: $[K]\rightarrow [K]$ over a size-$K$ set (\emph{e.g.}, actions) $\{a_0, \dots, a_{K-1}\}$ is defined as following, by additionally conditioned on current state $s$:
\begin{align}
P(\sigma|s, a_0, a_1, \dots, a_{K-1}) = \Pi_{k=0}^{K-1}\frac{\exp{(r(s, a_{\sigma_k}))}}{\sum_{j=k}^{K-1}\exp{(r(s, a_{\sigma_j}))}}
\label{eq:pl}
\end{align}
which represents the preference function over $K$ actions $\{a_0, \dots, a_{K-1}\}$. $\sigma$ is a permutation/ranking function: $[K]\rightarrow [K]$. $\sigma_0$ is the most preferred action index given current state $s$. $r(s,a_{\sigma_k})$ is the score of the $\sigma_k$-th ranked action for state $s$.

For $K=2$, PL reduces to pairwise preference relationship, which is known as the \textit{Bardley-Terry-Luce} model~\cite{bradley1952rank}.

For example, if $K=2$ and assuming preference $\sigma=[1,0]$ on set $\{a_0, a_1\}$, by Eq.~\eqref{eq:pl} we have:
\begin{align}
    P(a_1 \succ a_0)&:= P(\sigma|s, a_0, a_1) \\
    &= \frac{\exp(r(s, a_1))}{\exp{(r(s, a_1))}+\exp{(r(s,a_0))}} \cdot \frac{\exp{(r(s,a_0))}}{\exp{(r(s, a_0))}}\\
    &= \frac{1}{1+\exp\big(r(s, a_0) - r(s, a_1)\big)} \\
    &= \sigma\big(r(s,a_1) - r(s, a_0)\big)   \quad (\text{ Sigmoid function }\sigma(x) = \frac{1}{1+\exp(-x)})
\end{align}
and the last equation is used as the loss term in Eq.~\eqref{eq:rm_loss} for training the reward model in the proposed method, with slight modifications from single state-action pairs to state-action sequences.

\section{Details of Bi-DexHands Tasks}
\label{app:tasks}
\subsection{Overview}
\paragraph{Action.}
Each Shadow Hand has five fingers with 24 minimum drive units, including four underdriven fingertip units (Finger Distal: \texttt{FF1}, \texttt{MF1}, \texttt{RF1} and \texttt{LF1}). There are 20 proactive driven units, so the action space for each Shadow Hand is of 20 dimensions, following the original environment settings~\cite{chen2022towards}.
The dual Shadow Hands have 40 dimensions of action space, $\mathcal{A}_\text{left hand}=\mathcal{A}_\text{right hand}=20$. Additionally, in some tasks (\emph{e.g.}, \texttt{Switch}) the base of each Shadow Hand is movable. This leads to another 6 DoF for the translation and rotation of each hand base in the world frame. Details of the action spaces for each task are provided in Appendix \ref{app:act_space}.

\paragraph{Observation.}
The observation space consists of three components: $(\mathcal{O}_{\text{left hand}}, \mathcal{O}_{\text{right hand}}, \mathcal{O}_{\text{task}})$, representing the state information for the left and right Shadow Hand, and the task-relevant information. The dimensions of the observation space for the tasks range from 414 to 446. More information about the observation spaces for each task is detailed in Appendix \ref{app:obs_space}.

\paragraph{Reward.}

The task-completion reward function applied in experiments follow the original Bi-DexHands environment~\cite{chen2022towards}. Different tasks have different specific rewards but follow the same design principles. The reward is a dense function of (1) hand positions to grasping points $d_{left}, d_{right}$, (2) the object translation and rotation errors from the target $d_{target}$, and (3) penalties on actions $f(\boldsymbol{a})$ for smoothing trajectories:
\begin{equation}
    r(\boldsymbol{s}, \boldsymbol{a})= c_0 + c_1 d_{left} + c_2 d_{right} + c_3 d_{target} + c_4 f(\boldsymbol{a}),
\end{equation}
where $c_0, c_1, c_2, c_3, c_4$ are adjustable constants for each task. Details are provided in Appendix \ref{app:reward}.

\subsection{Detailed Description of Action Space}
\label{app:act_space}
Each Shadow Hand has five fingers with 24 minimum drive units, including four underdriven fingertip units. The thumb has 5 joints and 5 degrees of freedom, and all other fingers have 3 degrees of freedom and 4 joints. There are 20 proactive driven units (without Finger Distal: \texttt{FF1}, \texttt{MF1}, \texttt{RF1} and \texttt{LF1}), so the action space for each Shadow Hand is of 20 dimensions, $\mathcal{A}_\text{left hand}=\mathcal{A}_\text{right hand}=20$. At each step in simulation, the actions of the Shadow Hand are the absolute values of each joint angle as set points, and proportional-differential controller is used for low-level control. Some tasks have additionally 6 DoF for translation and rotation of the base for each hand, $\mathcal{A}_\text{left base}=\mathcal{A}_\text{right base}=6$. We summarize the complete action space for each task in Tab.~\ref{tab:act_space}. Only $\mathcal{A}_\text{left hand}$ and $\mathcal{A}_\text{right hand}$ are used for the RM, which are marked \textbf{bold} in Tab.~\ref{tab:act_space}. This design choice allows the RM to be applied over each hand separately without consideration of the base movement, thus the RM will focus only on the in-hand joint correlations. 

\begin{table*}[t]
\centering
\caption{Dimension of the action space in Bi-DexHands environment.}
\label{tab:act_space}
\resizebox{\columnwidth}{!}{ 
\begin{tabular}{c|ccc}
\toprule
      Task & \multirow{1}{*}{$\boldsymbol{\mathcal{A}_\text{left hand}}$=$\boldsymbol{\mathcal{A}_\text{right hand}}$} & \multirow{1}{*}{$\mathcal{A}_\text{left base}$=$\mathcal{A}_\text{right base}$} & \multirow{1}{*}{$\mathcal{A}_\text{left hand}+\mathcal{A}_\text{left base}+ \mathcal{A}_\text{right hand}+\mathcal{A}_\text{right base}$} \\ \hline
     \texttt{ShadowHand} (one hand) &  $\mathbf{20}$  & $0$     &  $20$    \\ \hline
     \texttt{Switch}  &  $\mathbf{20}$  & $6$      &  $52$  \\\hline
     \texttt{CatchOver2Underarm}  &  $\mathbf{20}$  & $6$      &  $52$    \\ \hline
     \texttt{CatchAbreast}  &  $\mathbf{20}$  & $6$      &  $52$  \\ \hline
     \texttt{HandOver}  &  $\mathbf{20}$   & $0$     &  $40$  \\ \hline
     \texttt{BlockStack} &  $\mathbf{20}$  & $6$      &  $52$ \\ \hline
     \texttt{CatchUnderarm}  &  $\mathbf{20}$   & $6$     &  $52$   \\ \hline
     \texttt{BottleCap} &  $\mathbf{20}$    & $6$    &  $52$  \\ \hline
     \texttt{LiftUnderarm} &  $\mathbf{20}$   & $6$     &  $52$  \\ \hline
     \texttt{TwoCatchUnderarm} &  $\mathbf{20}$   & $6$     &  $52$   \\ \hline
     \texttt{DoorOpenInward}  &  $\mathbf{20}$   & $6$     &  $52$  \\ \hline
     \texttt{DoorOpenOutward} &  $\mathbf{20}$   & $6$     &  $52$   \\ \hline
     \texttt{DoorCloseInward} &  $\mathbf{20}$  & $6$     &  $52$  \\ \hline
     \texttt{PushBlock}  &  $\mathbf{20}$   & $6$     &  $52$  \\ \hline
     \texttt{Scissors} &  $\mathbf{20}$  & $6$      &  $52$  \\ \hline
     \texttt{Pen} & $\mathbf{20}$  & $6$      &  $52$  \\ \hline
     \texttt{GraspAndPlace}  &  $\mathbf{20}$   & $6$     &  $52$   \\ \hline
     \texttt{Kettle}  &  $\mathbf{20}$  & $6$      &  $52$  \\ \hline
     \texttt{DoorCloseOutward} & $\mathbf{20}$  & $6$     &  $52$   \\ \hline
     \texttt{SwingCup}  &  $\mathbf{20}$  & $6$    &  $52$  \\ \midrule
\end{tabular}
}
\end{table*}

\subsection{Detailed Descriptions of Observation Space}
\label{app:obs_space}
For all tasks in the Bi-DexHands~\cite{chen2022towards} environment, the observation space consists of three parts: 
\begin{align}
    (\mathcal{O}_{\text{left hand}}, ~~\mathcal{O}_{\text{right hand}}, ~~\mathcal{O}_{\text{task}}),
\end{align}
where $\mathcal{O}_{\text{left hand}}$ and $\mathcal{O}_{\text{right hand}}$ are the state information of left and right Shadow Hands, respectively. $\mathcal{O}_{\text{task}}$ represents the task-relevant information. $\mathcal{O}_{\text{left hand}}$ and $\mathcal{O}_{\text{right hand}}$ are symmetric in all tasks due to the same structure description file, therefore they are made of the same components, which are detailed as following:


\begin{itemize}[leftmargin=*]
\item $\mathcal{J}_{p}, \mathcal{J}_{v}, \mathcal{J}_{f}\in \mathbb{R}^{24}$, corresponds to all joint degree of freedom of angle, velocity, and force with drive units, respectively.
\item $\mathcal{P}_w, \mathcal{R}_w \in \mathbb{R}^{3}$ represents the position and rotation of the base of the hand.
\item $\mathcal{FT}_i = [FT_{\text{pose}}, FT_{v_l}, FT_{v_a}, FT_{f}, FT_{t}] \in \mathbb{R}^{19\times5}$, corresponds to the pose, linear velocity, angular velocity, force magnitude, and torque of each fingertip, respectively.
\item $\mathcal{A} \in \mathbb{R}^{20/26}$, indicates the action executed by the hand in the previous step, which is consistent with the action space.
\end{itemize}

With the above definitions, the state information of one Shadow Hand can be represented as $\mathcal{O}_{\text{left hand}} = \mathcal{O}_{\text{right hand}} = \{\mathcal{J}_{p}, \mathcal{J}_{v}, \mathcal{J}_{f}, \mathcal{P}_w, \mathcal{R}_w, \{\mathcal{FT}_i\}_{i=1}^{5}, \mathcal{A}\}$. 


The task-relevant information $\mathcal{O}_{\text{task}}$ apart from hands includes state information of the manipulated objects, etc. $\mathcal{O}_{\text{task}}$ is task-specific. We summarize the complete observation space for each task in Tab.~\ref{tab:obs_space}. Only $\mathcal{J}_p$ in $\mathcal{O}_\text{left hand}$ and $\mathcal{O}_\text{right hand}$ are used for the RM, which are marked \textbf{bold} in Tab.~\ref{tab:obs_space}. These are also in-hand state information only and separable for two hands.

\begin{table*}[t]
\caption{Dimension of the observation space in Bi-DexHands environment.}
\label{tab:obs_space}
\resizebox{\columnwidth}{!}{ 
\begin{tabular}{cccccccccc}
\toprule
        \multirow{2}{*}{Task} &  \multicolumn{7}{c}{\multirow{1}{*}{$\mathcal{O}_\text{left hand}$=$\mathcal{O}_\text{right hand}$}} &  \multirow{2}{*}{$\mathcal{O}_\text{task}$} &  \multirow{2}{*}{$\mathcal{O}_\text{left hand}+\mathcal{O}_\text{right hand}+\mathcal{O}_\text{task}$} \\ \cline{2-8}
      & $\boldsymbol{\mathcal{J}_p}$   &  $\mathcal{J}_v$  & $\mathcal{J}_f$ & $\mathcal{P}_w$ & $\mathcal{R}_w$ & $\{\mathcal{FT}_i\}_{i=1}^{5}$ & $\mathcal{A}$ \\ \midrule
     \texttt{ShadowHand} (one hand)  &  $\mathbf{24}$        &  $24$  & $24$ &  $0$  &   $0$  &  $95$  &  $20$  &  $24$ &  $211$\\ \hline
     \texttt{Switch}  &  $\mathbf{24}$        &  $24$  & $24$ &  $3$  &   $3$  &  $95$  &  $26$  &  $30$ &  $428$\\\hline
     \texttt{CatchOver2Underarm}  &  $\mathbf{24}$        &  $24$  & $24$ &  $3$  &   $3$  &  $95$  &  $26$  &  $24$ &  $422$\\ \hline
     \texttt{CatchAbreast}  &  $\mathbf{24}$        &  $24$  & $24$ &  $3$  &   $3$  &  $95$  &  $26$  &  $24$ &  $422$\\ \hline
     \texttt{HandOver}  &  $\mathbf{24}$        &  $24$  & $24$ &  $0$  &   $0$  &  $95$  &  $20$  &  $24$ &  $398$  \\ \hline
     \texttt{BlockStack}  &  $\mathbf{24}$        &  $24$  & $24$ &  $3$  &   $3$  &  $95$  &  $26$  &  $30$ &  $428$\\ \hline
     \texttt{CatchUnderarm}  &  $\mathbf{24}$        &  $24$  & $24$ &  $3$  &   $3$  &  $95$  &  $26$  &  $24$ &  $422$\\ \hline
     \texttt{BottleCap}  &  $\mathbf{24}$        &  $24$  & $24$ &  $3$  &   $3$  &  $95$  &  $26$  &  $16$ &  $414$\\ \hline
     \texttt{LiftUnderarm}  &  $\mathbf{24}$        &  $24$  & $24$ &  $3$  &   $3$  &  $95$  &  $26$  &  $30$ &  $428$\\ \hline
     \texttt{TwoCatchUnderarm}  &  $\mathbf{24}$        &  $24$  & $24$ &  $3$  &   $3$  &  $95$  &  $26$  &  $48$ &  $446$\\ \hline
     \texttt{DoorOpenInward}  &  $\mathbf{24}$        &  $24$  & $24$ &  $3$  &   $3$  &  $95$  &  $26$  &  $30$ &  $428$\\ \hline
     \texttt{DoorOpenOutward}&  $\mathbf{24}$        &  $24$  & $24$ &  $3$  &   $3$  &  $95$  &  $26$  &  $30$ &  $428$\\ \hline
     \texttt{DoorCloseInward} &  $\mathbf{24}$        &  $24$  & $24$ &  $3$  &   $3$  &  $95$  &  $26$  &  $30$ &  $428$ \\ \hline
     \texttt{PushBlock}  &  $\mathbf{24}$        &  $24$  & $24$ &  $3$  &   $3$  &  $95$  &  $26$  &  $30$ &  $428$\\ \hline
     \texttt{Scissors}  &  $\mathbf{24}$        &  $24$  & $24$ &  $3$  &   $3$  &  $95$  &  $26$  &  $30$ &  $428$\\ \hline
     \texttt{Pen} &  $\mathbf{24}$        &  $24$  & $24$ &  $3$  &   $3$  &  $95$  &  $26$  &  $30$ &  $428$\\ \hline
     \texttt{GraspAndPlace} &  $\mathbf{24}$        &  $24$  & $24$ &  $3$  &   $3$  &  $95$  &  $26$  &  $30$ &  $428$\\ \hline
     \texttt{Kettle} &  $\mathbf{24}$        &  $24$  & $24$ &  $3$  &   $3$  &  $95$  &  $26$  &  $30$ &  $428$\\ \hline
     \texttt{DoorCloseOutward}  &  $\mathbf{24}$        &  $24$  & $24$ &  $3$  &   $3$  &  $95$  &  $26$  &  $30$ &  $428$\\ \hline
     \texttt{SwingCup} &  $\mathbf{24}$        &  $24$  & $24$ &  $3$  &   $3$  &  $95$  &  $26$  &  $30$ &  $428$\\ \bottomrule
\end{tabular}
}
\end{table*}

\subsection{Details of Task Reward Function}
\label{app:reward}
Here we describe the terms $d_{left}, d_{right}, d_{target}, f(\boldsymbol{a})$ in the reward function.
Denote the object and goal positions as $x_o$ and $x_g$ respectively. Then, the translational position difference between the object and the goal $d_t$ is given by $d_t=\Vert x_o-x_g \Vert_2$. Denote the angular position difference between the object and the goal as $d_a$, then the rotational difference $d_r$ is given by $d_r = 2\arcsin{\text{clamp}(\Vert d_a \Vert_2, \text{max }= 1.0)}$. For object-catching tasks including \texttt{HandOver}, \texttt{CatchUnderarm}, \texttt{CatchOver2Underarm}, \texttt{CatchAbreast} and \texttt{TwoCatchUnderarm}, the reward is just:
\begin{equation}
    r = d_{target}+c_4 f(\boldsymbol{a}) = exp[-0.2(\alpha d_t + d_r)] + c_4 f(\boldsymbol{a})
\end{equation}
where $\alpha$ is a constant balancing translational and rotational rewards. $f(\boldsymbol{a})=\sum_{i\in[|\mathcal{A}|]} a_i^2$ is the action penalty term. 

Other tasks may have a different $d_{target}$ and use hand position match terms $d_{left}, d_{right}$. The position difference between the left hand to the left handle $d_{left}$ is given by $d_{left}=\Vert x_{lhand}-x_{lhandle} \Vert_2$.The position difference between the right hand to the right handle $d_{right}$ is given by $d_{right}=\Vert x_{rhand}-x_{rhandle} \Vert_2$. The reward is then given by:
\begin{equation}
        r= c_0 + c_1 d_{left} + c_2 d_{right} + c_3 d_{target} + c_4 f(\boldsymbol{a})
\end{equation}
For the specific reward and constant values for each task, we refer the reader to the paper \cite{chen2022towards}.

\section{Additional Details in Policy Generation}
\subsection{Policy Diversity}
\label{app:diversity}
\begin{figure}[htbp]
	\centering\includegraphics[width=0.8\columnwidth]{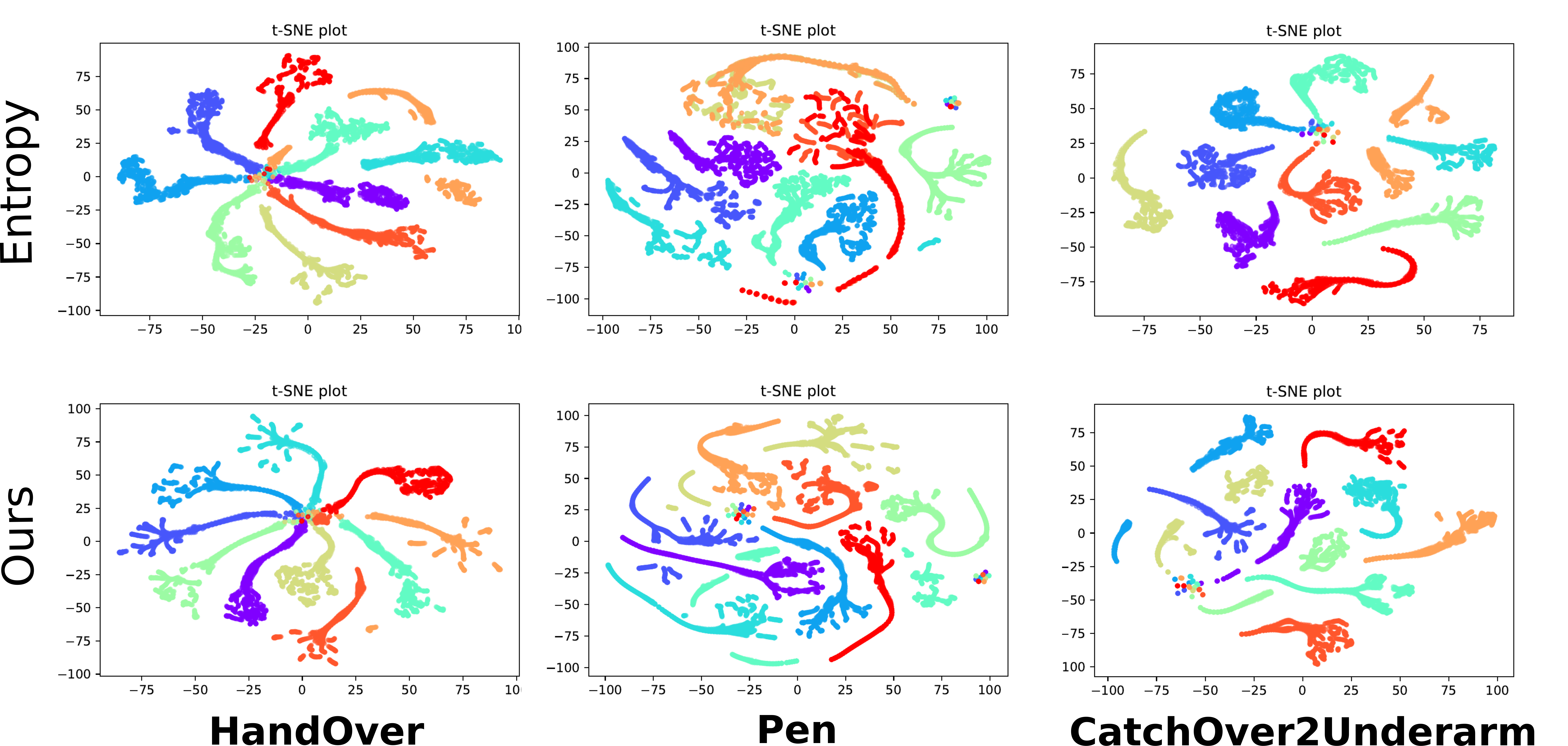}
	\caption{Visualization of t-SNE plots for policies trained with diverse policy generation methods on three tasks. Each color corresponds to the one policy.}
	\label{fig:tsne_compare}
\end{figure}

\paragraph{T-SNE Plots.}
To demonstrate the diversity of the polices trained with Alg.~\ref{alg:diverse_ppo}, we display the t-SNE plots on the generated polices. The observations along trajectories for the trained policies on three tasks (\texttt{HandOver}, \texttt{Pen}, \texttt{CatchOver2Underarm}) are visualized in Fig.~\ref{fig:tsne_compare} in a projected space. The top row is the PPO policies with entropy regularization, and the bottom row is our method with diversity loss Eq.~\eqref{eq:add_loss}. For each trained policy, 10 trajectories are collected by deploying the policy in the corresponding environment. The t-SNE~\cite{van2008visualizing} plots show the clustering results in embedded 2D space for observed states in collected trajectories with different policies. The observed states here only involve the joint positions of the two hands, since other observations like velocities or forces may have different magnitudes of values and unfairly affect the results. 

\paragraph{Quantitative Comparison.}
 We also quantitatively evaluate the diversity of generated policies in comparison of different methods. Three methods are compared: (1) PPO without any diversity bonus (\textbf{No Bonus}); (2) PPO with the policy entropy bonus (\textbf{Entropy}); (3) our proposed PPO with diversity loss Eq.~\eqref{eq:add_loss} and Alg.~\ref{alg:diverse_ppo} for generation (\textbf{Ours}).
For the policy entropy approach, the original PPO loss introduced in Sec.~\ref{sec:prim} are modified to be: $\mathcal{J}(\pi_\theta; r)=-\mathbb{E}_{s\sim \rho_\pi, a\sim\pi}[\min(R_\theta A(s,a), \text{clip}(R_\theta, 1-\epsilon, 1+\epsilon)A(s,a))-\zeta \mathcal{H}(\pi(a|s))]$, where $\mathcal{H}(\cdot)$ is the entropy and $\zeta=0.02$ in the experiments.

For each method, 10 policies are trained for 20000 episodes under different random seeds on three tasks \texttt{HandOver}, \texttt{Pen} and \texttt{CatchOver2Underarm}. These three tasks generally have higher success rates compared with other tasks, and we believe they are more reasonable for conducting further evaluation study. For each trained policy, 10 trajectories are collected by depolying the policy in the corresponding environment. The pairwise Euclidean distance of trajectories are compared across all policies trained with each of three methods to quantitatively evaluating the effectiveness of the proposed diverse policy generation procedure as introduced in Sec.~\ref{sec:diverse}. Both the success rates and policy diversity (trajectory Euclidean distance) are summarized in Table~\ref{tab:diversity_compare2}. As shown in the results, the proposed method for diverse policy generation achieve the highest success rates in two of three tasks and highest diversity values for all three tasks. This testifies the effectiveness of the proposed method for providing diverse policies for downstream tasks, as well as maintaining high success rates for the task completion.

\begin{table*}[t]
\centering
\caption{Diversity and success rate of three methods for policy training on three tasks.}
\label{tab:diversity_compare2}  
\begin{tabular}{c|ccc|ccc}
\toprule
       &  \multicolumn{3}{c|}{Success Rate} & \multicolumn{3}{c}{Diversity} \\ \hline
      \backslashbox{Task}{Method} & No Bonus & Entropy & Ours & No Bonus & Entropy & Ours \\ \hline
     \texttt{HandOver}  &  $\mathbf{85.4\pm 26.5}$   & $77.3\pm 23.3$     &  $85.2\pm 26.3$ & $53.0$ & $55.0$ & $\mathbf{58.7}$ \\ \hline
     \texttt{Pen} & $74.8\pm 38.9$  & $70.4\pm 36.7$      &  $\mathbf{89.3\pm 27.7}$ & $77.9$  & $75.9$ & $\mathbf{78.4}$  \\ \hline
     \texttt{CatchOver2Underarm}  &  $76.2\pm 23.5$ & $65.4\pm 26.7$      &  $\mathbf{81.3\pm 24.5}$  & $53.8$ & $54.7$ & $\mathbf{57.3}$   \\ \midrule
\end{tabular}
\end{table*}

\subsection{Trajectory Length}
\label{app:traj_length}
\begin{figure}[htbp]
	\centering\includegraphics[width=7.5 cm]{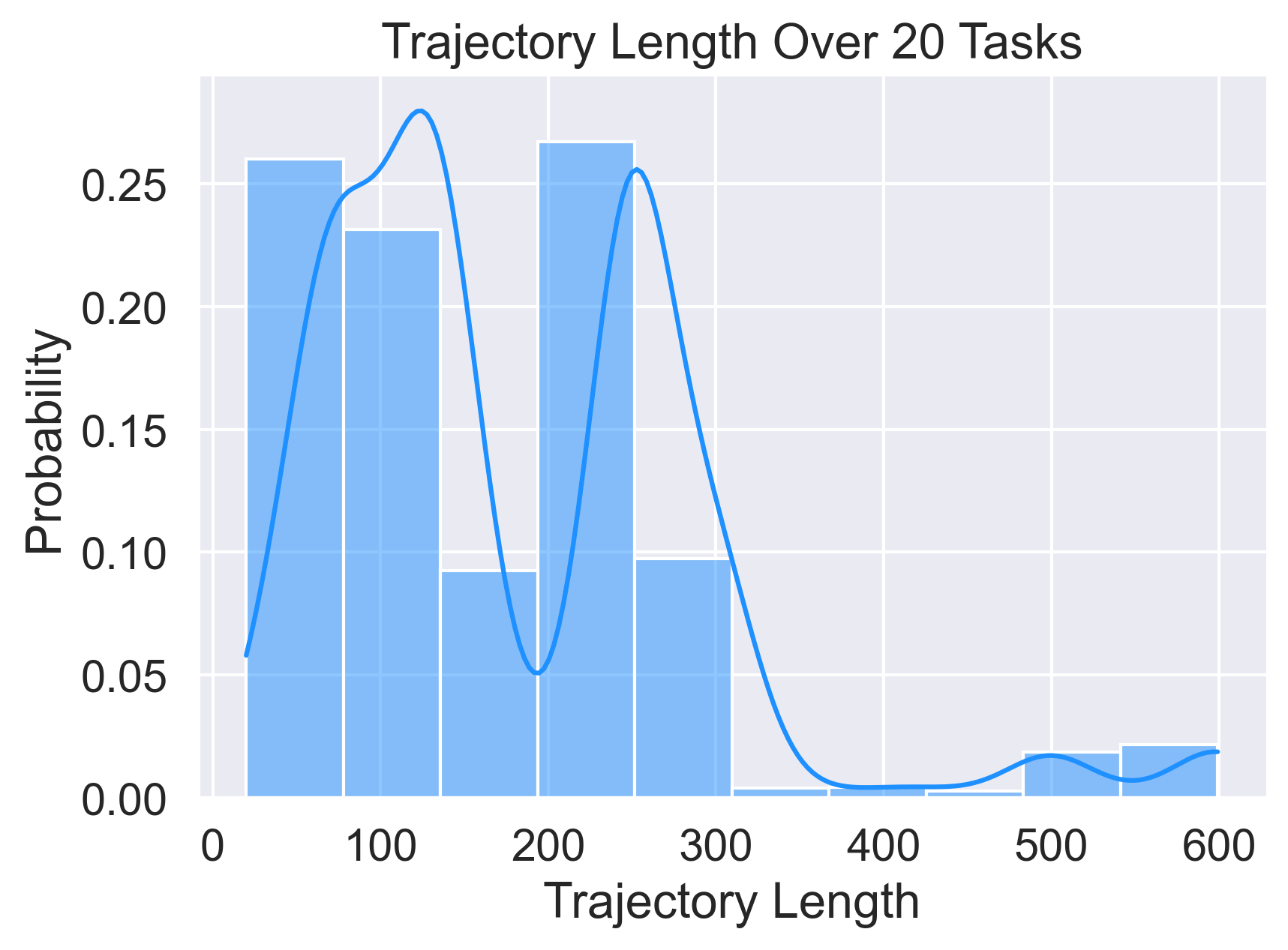}
	\caption{Histogram of trajectory length for the 20 tasks. }
	\label{fig:all_len}
\end{figure}

\begin{figure*}[htbp]   
	\centering\includegraphics[width=\textwidth]{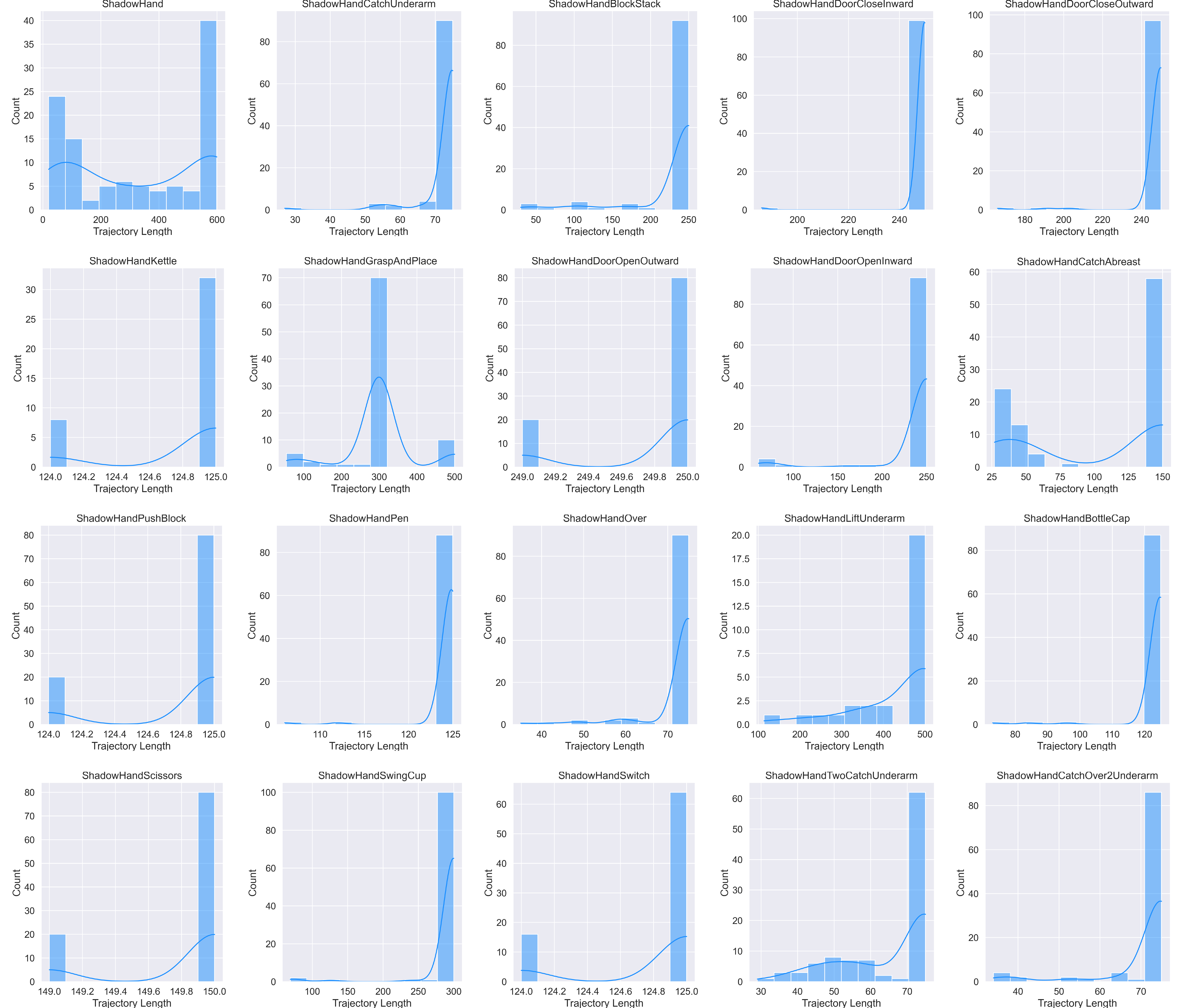}
    	\caption{The distribution of trajectory length for 20 tasks in Bi-DexHands.}
	\label{fig:traj_len}
\end{figure*}

We show the distributions of the trajectory length executed by the trained policies over the 20 tasks, as in Fig.~\ref{fig:traj_len}. The results are further summarized in Fig.~\ref{fig:all_len}. Each distribution is derived with 10 random seeds of policies after training. 

\subsection{Learning Curves}
\label{app:learning_curve}
\begin{figure*}[htbp]   
	\centering\includegraphics[width=\textwidth]{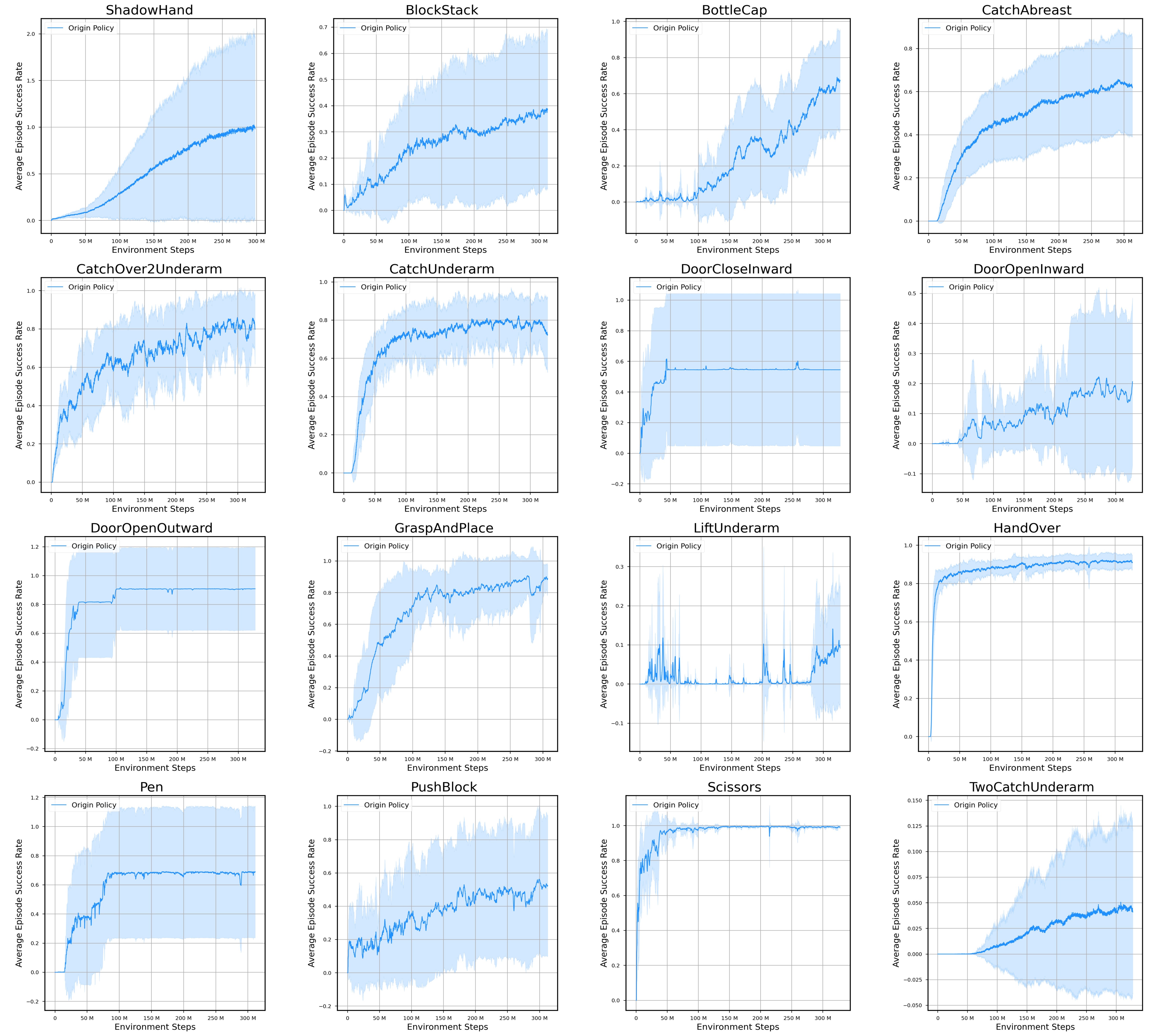}
    	\caption{The learning curves for policies without RM on 17 training tasks.}
	\label{fig:curve_seen}
\end{figure*}

\begin{figure*}[htbp]   
	\centering\includegraphics[width=\textwidth]{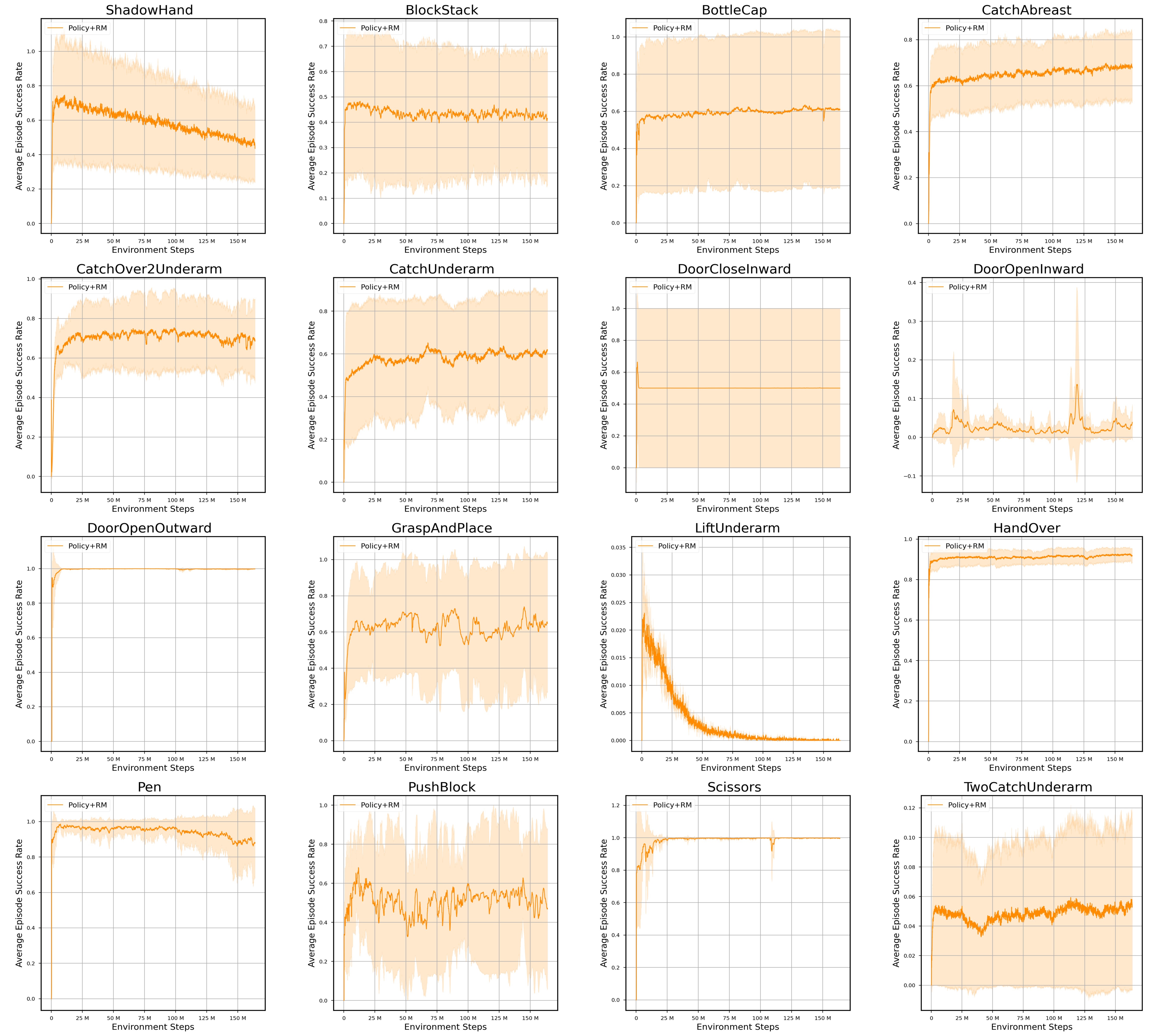}
    	\caption{The learning curves for policies with RM (fine-tuning) on 17 training tasks.}
	\label{fig:rm_curve_seen}
\end{figure*}

Fig.\ref{fig:curve_seen} show the learning curves for policies without RM on 17 training tasks. Fig.~\ref{fig:rm_curve_seen} show the learning curves for policies with RM (fine-tuning) on 17 training tasks.

\begin{figure*}[htbp]   
	\centering\includegraphics[width=\textwidth]{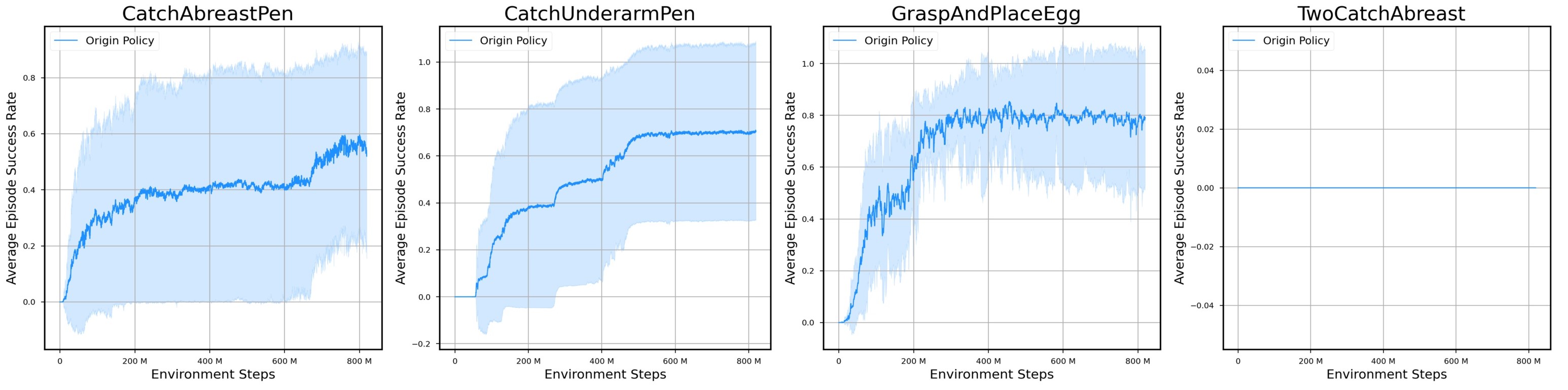}
    	\caption{The learning curves for policies without RM on 4 unseen tasks.}
	\label{fig:curve_unseen}
\end{figure*}

\begin{figure*}[htbp]   
 	\centering\includegraphics[width=\textwidth]{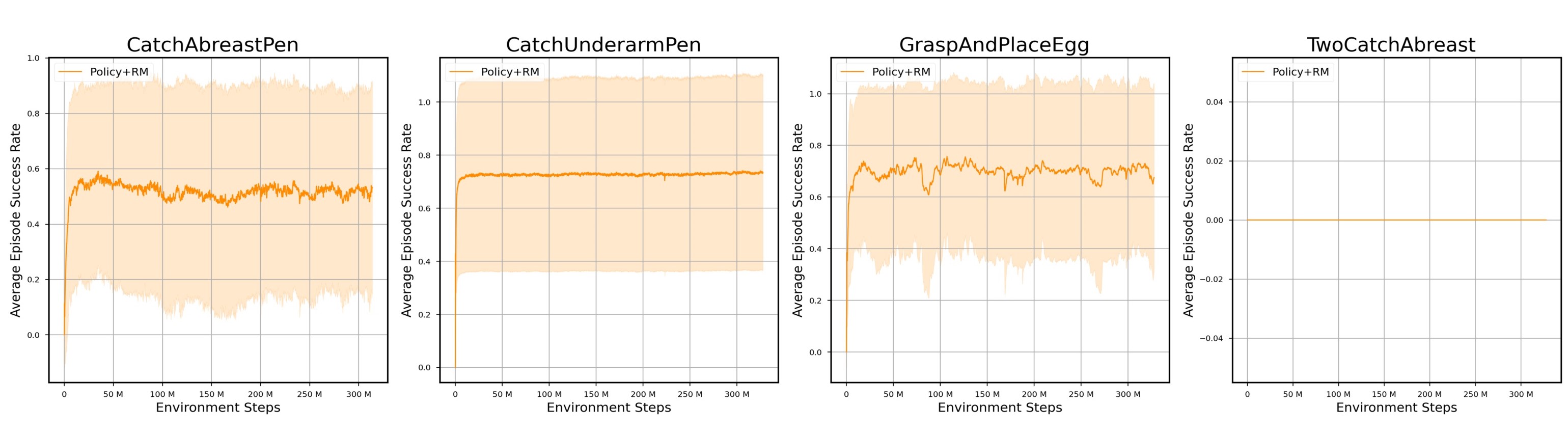}
    	\caption{The learning curves for policies with RM (fine-tuning) on 4 unseen tasks.}
	\label{fig:rm_curve_unseen}
\end{figure*}

Fig.\ref{fig:curve_unseen} show the learning curves for policies without RM on 4 unseen tasks. Fig.~\ref{fig:rm_curve_unseen} show the learning curves for policies with RM (fine-tuning) on 4 unseen tasks. 

\newpage
\section{Comparison with Reward Engineering}
\label{app:compare_re}
We compare the applied RLHF method with the reward engineering (RE) approach, in which the reward function is augmented by adding penalty terms for achieving smooth trajectories, as in previous work~\cite{sievers2022learning, qi2023hand}. Specifically, following the work~\cite{qi2023hand}, the added penalties include the torque penalty and the energy consumption penalty, and the coefficients are carefully chosen to match with the reward scale in each task. The penalty term is written as:
\begin{align}
    r_\text{Penalty} = c_5||\boldsymbol{\tau}||^2 + c_6\boldsymbol{\tau}^\intercal \boldsymbol{\dot{q}}
\end{align}
where $||\boldsymbol{\tau}||^2$ is the torque penalty on all joints, and $\boldsymbol{\tau}^\intercal \boldsymbol{\dot{q}}$ is the energy consumption penalty, with $\boldsymbol{\dot{q}}$ indicating the pose changes. The coefficients $c_5, c_6$ are chosen to be $0.01, 0.02$ respectively multiplied by the scale of task completion reward to achieve adaptive adjustment of penalty values for different tasks. Finally, for the `Policy+RE' approach as in Table.~\ref{tab:compare_re}, the policies are trained with the reward: $r=r_\text{Task}+r_\text{Penalty}$.

\begin{figure*}[htbp]   
 	\centering\includegraphics[width=\textwidth]{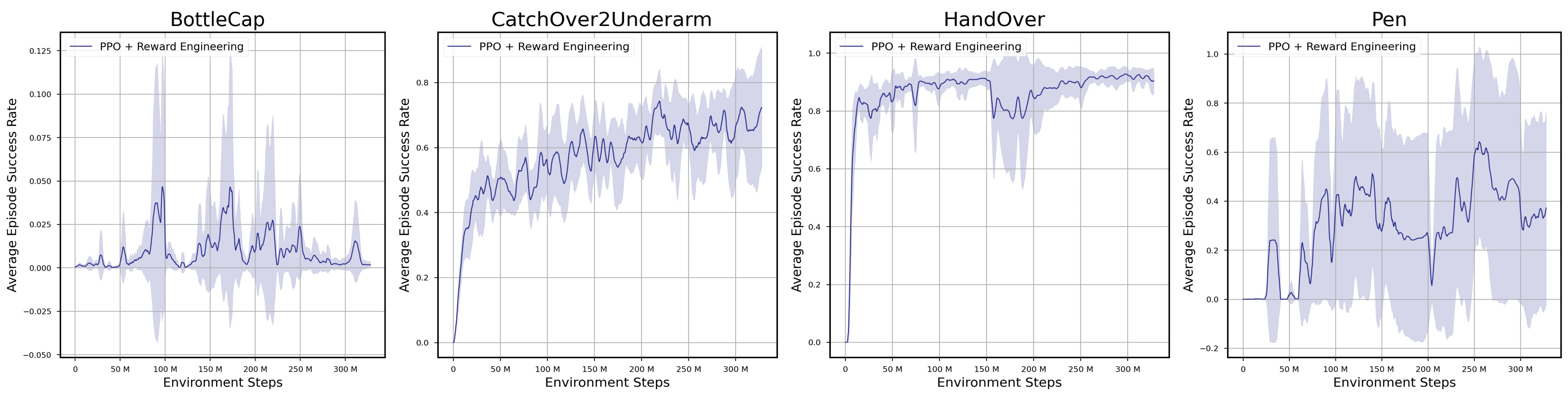}
    	\caption{The learning curves for policies with reward engineering on 4 tasks.}
	\label{fig:re_curve}
\end{figure*}

We compare the proposed `Policy+RM' with `Policy+RE' approaches on four tasks \texttt{CatchOver2Underarm}, \texttt{HandOver}, \texttt{BottleCap} and \texttt{Pen}, each trained with 5 random seeds for 20000 episodes. The resulting success rates and learning curves are shown in Table.~\ref{tab:compare_re} and Fig.~\ref{fig:re_curve}. It can be shown that by adding penalty terms in the reward function, the task completion performances can be easily destroyed, leading to a $26\%$ drop compared from original polices and $24\%$ gap from our proposed method on average.

\begin{figure*}[htbp]   
 	\centering\includegraphics[width=\textwidth]{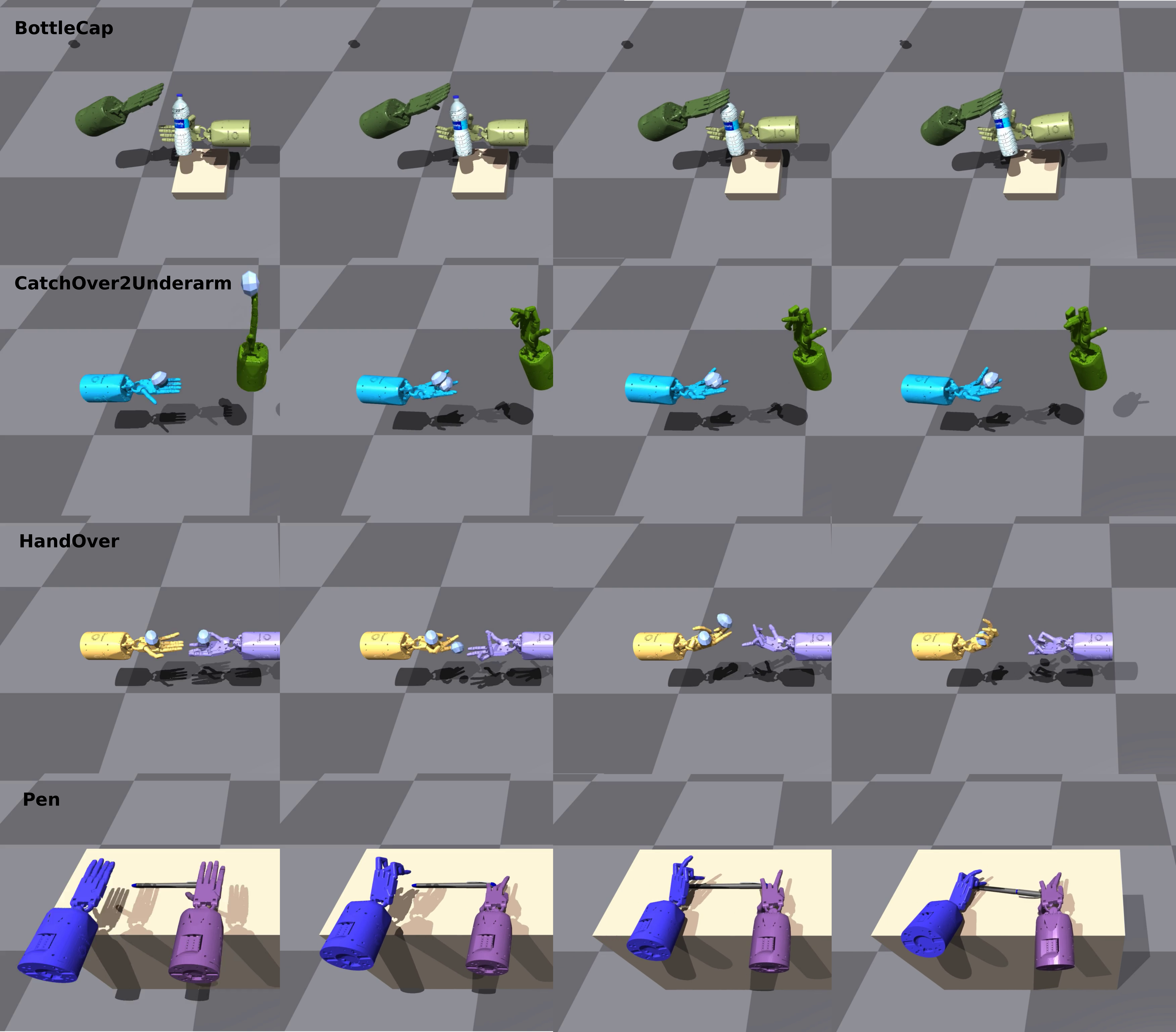}
    	\caption{Visualization of hand and object trajectories in four tasks for polices trained with reward engineering (RE). From left to right it shows the procedure of the task completion.}
	\label{fig:re_vis}
\end{figure*}

In terms of human-like behaviors, we find polices trained with RE display unnatural behaviors as shown in Fig.~\ref{fig:re_vis}. Although the torque and energy penalties indeed regulate the behaviors to display less unnecessary motion on hands, it tend to make the joints on hands stiffer. This can be obvious in the task \texttt{BottleCap}, where the fingers on hand for twisting the cap are almost straight all the time and significantly hurt the task completion. For other tasks, we also show that adding penalty terms are not sufficient to generate human-like behaviors. For example, the fingers are unnaturally twisted in \texttt{CatchOver2Underarm}, and some fingers are stretching in a weird way in \texttt{HandOver} and \texttt{Pen}. All these observations verify that the human-likeliness requires some more sophisticated regularization on policies rather than just adding penalty terms. 


\section{Reward Model Evaluation}
\label{app:rm_eval}
The reward model needs to (1) be consistent with labelers preference and (2) provide information gain. (1) can be evaluated on whether the RM matches with the labelers' preference, and this relationship can be task specific. Some tasks like \texttt{SwingCup} may not show the consistence, and therefore being discarded in later training and feedback collection. (1) can be measured by evaluating the RM score on those policies, and comparing it with the human preferences. Assuming (1) holds, sastifying (2) means RM needs to have a clear preference over the existing policies for each task. According to these principles, the Eq.~\eqref{eq:human_prefer_score} is used in RM evaluation experiments.

\subsection{Consistency between RM and Human Preference}
Before fine-tuning the task policies, we evaluate the RM from human preferences, by letting the labelers to provide the human preference scores as comparison with the average rewards from the RM over the comparing trajectories. The results are displayed in Fig.~\ref{fig:rm_compare} for two tasks. The ten sets of polices with random seeds are evaluated in this experiment, 25 trajectories are collected for each policy set. The top two figures show the evaluated scores with the RM, averaged over the entire trajectories. The two figures at the bottom show the human preference score $c_\text{HF}$ as Eq.~\eqref{eq:human_prefer_score} by directly collecting human preferences over randomly paired trajectories for different policy sets. 

The results in Fig.~\ref{fig:rm_compare} show the consistency of RM scores and the human preference scores. For each task, the human preference scores are evaluated on a randomly sampled batch of trajectories for collecting human preferences, and the RM score is evaluated on the whole dataset collected with the policy sets. As a result, there might be slight differences between the human preference score and the RM scores for some model indices. In general, the trained RM represents the preference bias from human feedback. The breakdown results for each labelers are discussed in the next section.

\begin{figure}[htbp]
\centering\includegraphics[width=0.6\textwidth]{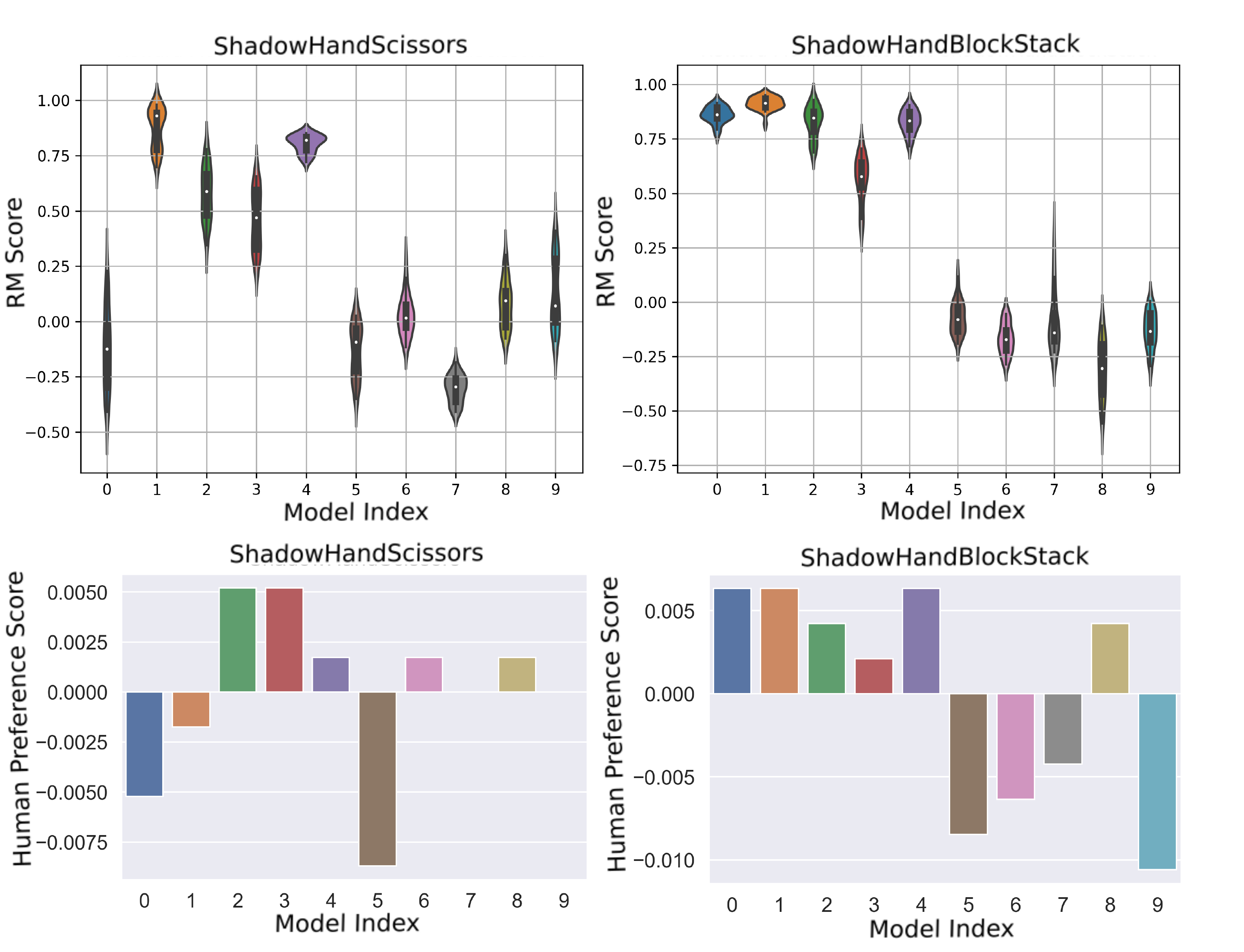}
	\caption{Comparison of the accumulative rewards over trajectories by RM (top) and human preference scores (bottom). }
	\label{fig:rm_compare}
\vskip -0.1in
\end{figure}

\subsection{Breakdown of Human Preference Results} 
\label{app:break_human_pre}
\begin{figure*}[htbp]
\centering\includegraphics[width=\textwidth]{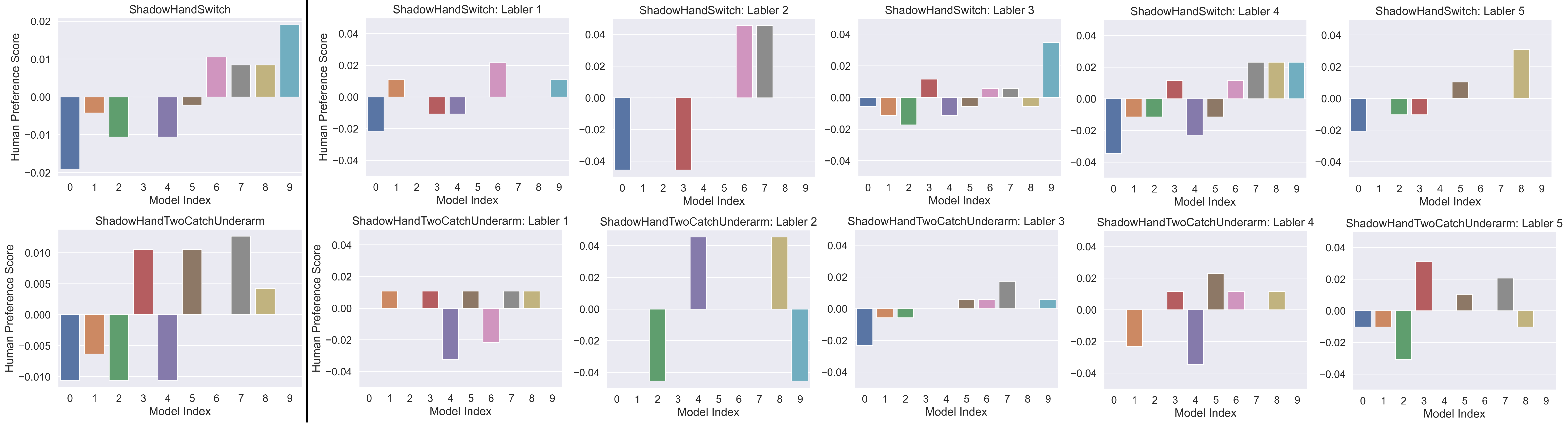}
	\caption{Human preference scores from different labelers on two tasks: \texttt{Switch} (top) and \texttt{TwoCatchUnderarm} (bottom). Left side is the summarized preference score over all labelers. Right side is the breakdown results for each labeler. }
	\label{fig:diff_label}
\end{figure*}
As an example, the labeled results for two tasks and the breakdown of individual feedback from five labelers are displayed in Fig.\ref{fig:diff_label}. Since the paired trajectories are randomly sampled from the dataset for each labeler, different labelers may not receive the same distribution of data samples over the policy sets. Also Fig.\ref{fig:diff_label} shows the average preference over a policy set, but not the per-trajectory preference statistics, some missing bars on the diagrams indicate a neural preference score ($\sim 0$) over the policy set. A majority of the preference scores are aligned with the overall preference across labelers. However, the disagreement can also happen among different labelers.

\subsection{More RM Evaluation Results}
\begin{figure*}[!h]
	\centering\includegraphics[width=\textwidth]{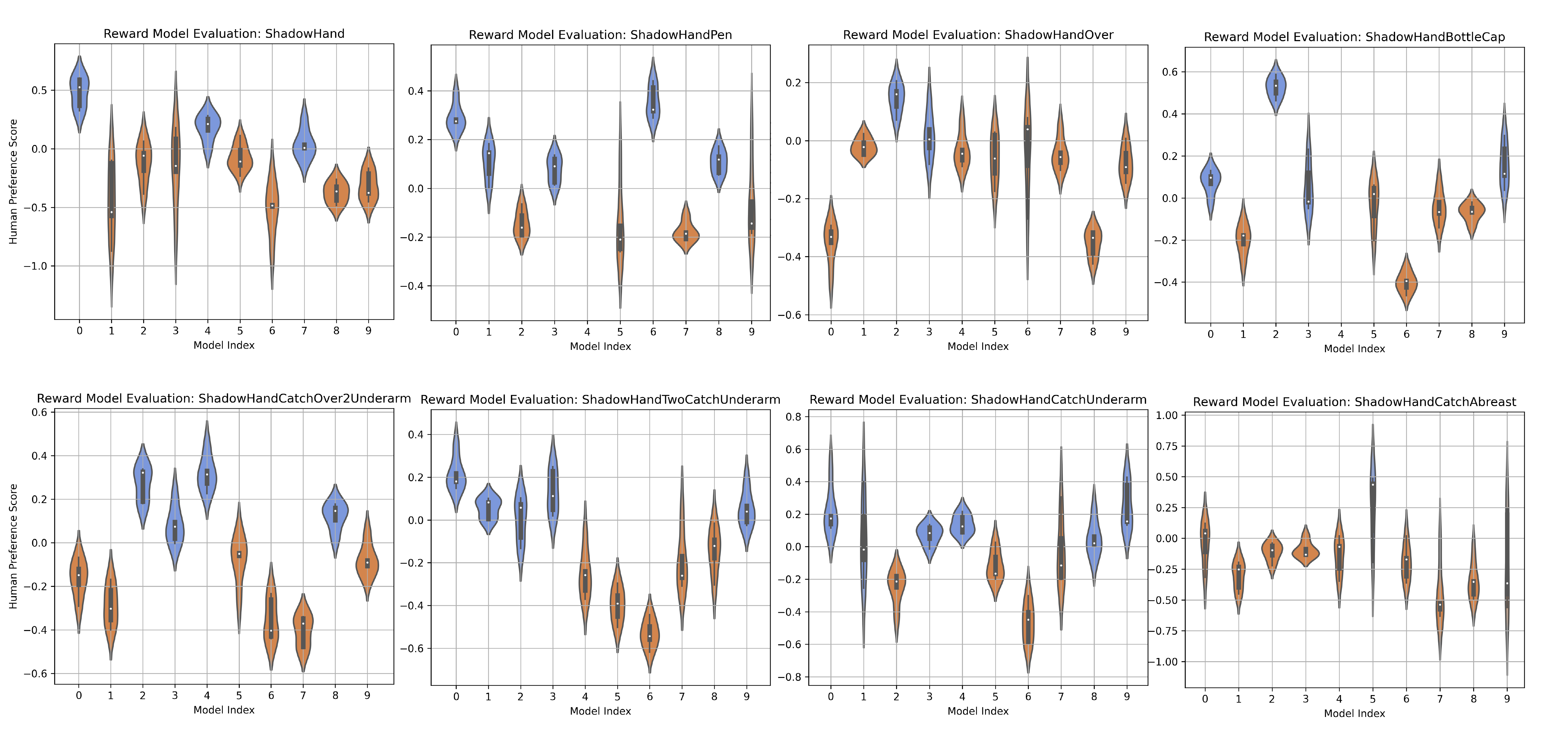}
    \centering\includegraphics[width=\textwidth]{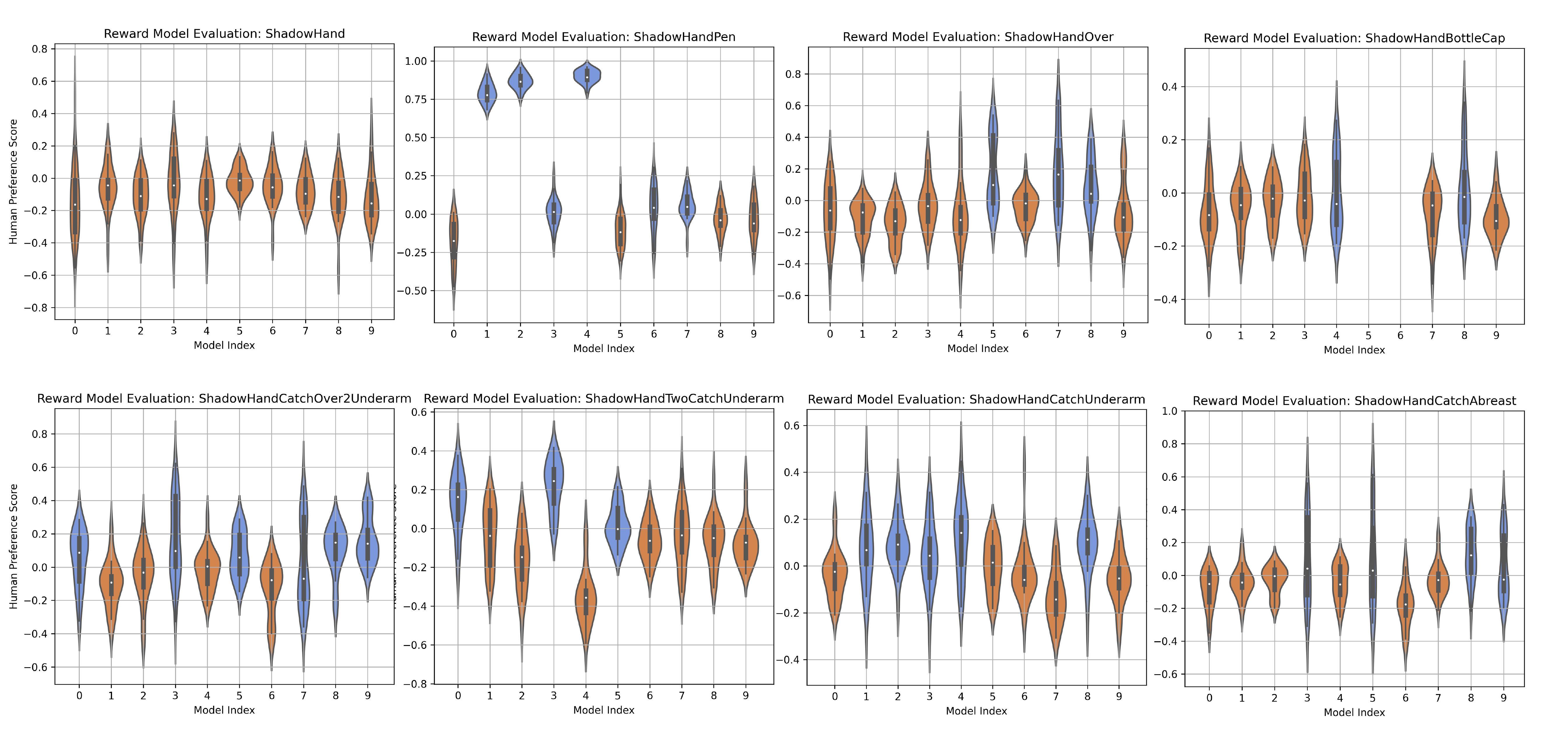}
    \centering\includegraphics[width=\textwidth]{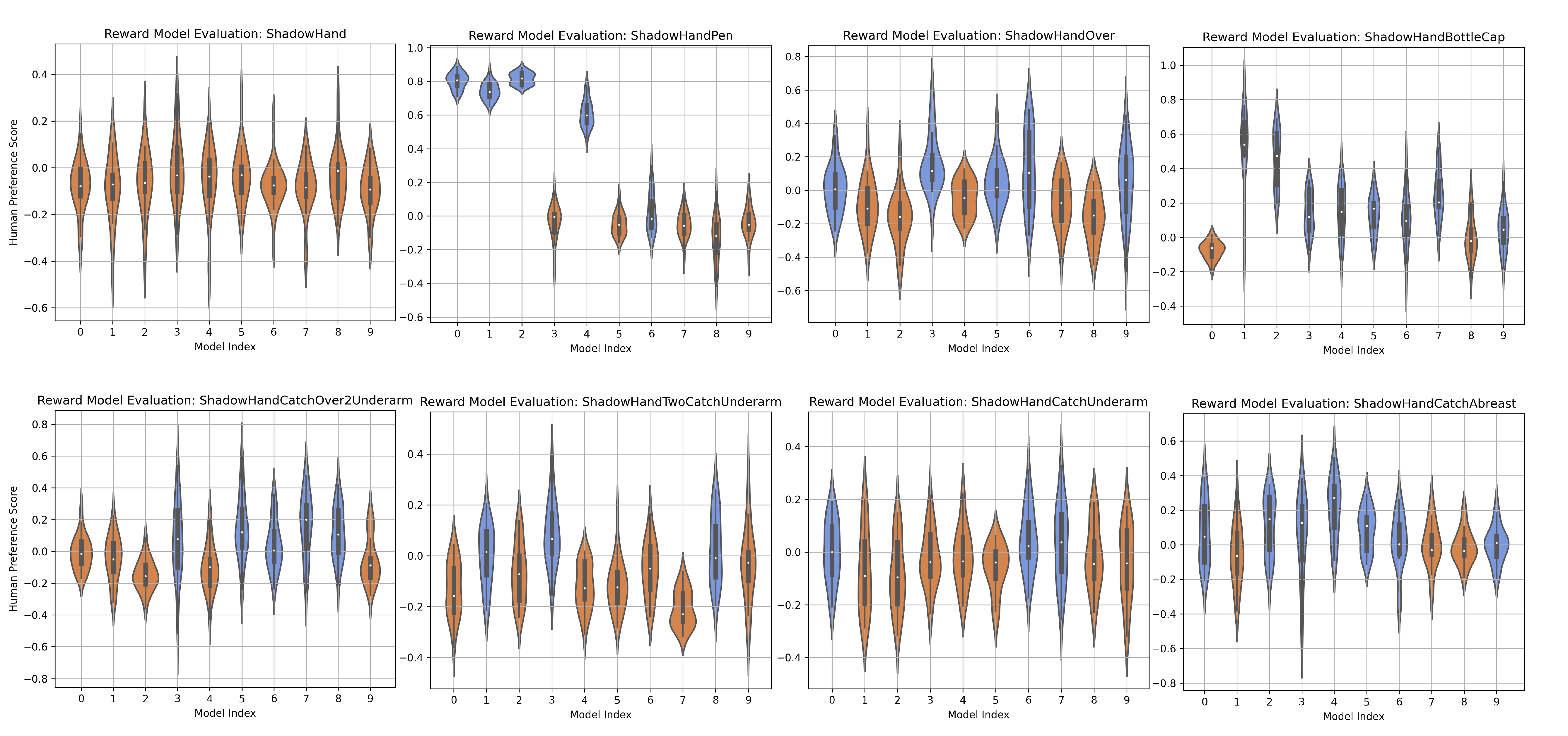}
	\caption{The RM evaluated over 8 tasks for different iterations. Top 2 rows: first iteration; Middle 2 rows: second iteration; Bottom 2 rows: third iteration.}
	\label{fig:rm_eval}
\end{figure*}

Fig.~\ref{fig:rm_eval} shows the RMs evaluated on the trajectories for different policy sets in each iteration, over 8 tasks. The results show that as the number of iterations increases, the RM tends to provide more neutral preference results, which indicates a more delicate preference over the human-like behaviors.

\section{Hyperparameters}
\label{app:hyper}

Table~\ref{ppo_para} summarizes the hyperparameters for training task-specific polices with the PPO algorithm.
\begin{table*}[htbp]
    \centering
    \caption{Hyperparameters of PPO.}
    \resizebox{\columnwidth}{!}{
    \begin{tabular}{c|c|c|c|c}
    \toprule 
    Hyperparameters & Other Tasks & \texttt{LiftUnderarm} & \texttt{BlockStack} & \texttt{TwoCatchUnderarm}\\\hline
    Num mini-batches & 4 & 4 & 8 & 4\\
    Num opt-epochs & 5 & 10 & 2 & 5\\
    Num episode-length & 8 & 20 & 8 & 8\\\hline
    Hidden size & [1024, 1024, 512] & [1024, 1024, 512] & [1024, 1024, 512] & [1024, 1024, 512]\\
    Clip range & 0.2 & 0.2 & 0.2 & 0.2\\
    Max grad norm & 1 & 1 & 1 & 1\\
    Learning rate & 3.e-4 & 3.e-4 & 3.e-4 & 3.e-4\\
    Discount ($\gamma$) & 0.96 & 0.96 & 0.9 & 0.96\\
    GAE lambda ($\lambda$) & 0.95 & 0.95& 0.95 & 0.95\\
    Init noise std & 0.8 & 0.8& 0.8 & 0.8\\\hline
    Desired kl & 0.016 & 0.016 & 0.016 & 0.016\\
    Ent-coef & 0 & 0 & 0 & 0.02\\
    \bottomrule
    \end{tabular}
    }
    \label{ppo_para}
\end{table*}

\section{Detailed Preference Results}
\label{app:preference_breakdown}
Tab.~\ref{tab:prefer_breakdown} shows the breakdown results of preference for each environment for the first iteration. Tab.~\ref{tab:prefer_breakdown2} shows the breakdown results of preference for each environment for the final iteration.

\begin{table}[htbp]
\caption{Preference results for each training task (first iteration).}
\label{tab:prefer_breakdown}
\centering
\begin{tabular}{c|ccc}
\toprule
      Task &  \multirow{1}{*}{Policy+RM} & \multicolumn{1}{c}{Original Policy} & \multicolumn{1}{c}{Not Sure} \\ \hline
     \texttt{ShadowHand}  &  $0.257$        &  $0.285$   &  $0.457 $ \\ \hline
     \texttt{Switch}  &  $0.272$        &  $0.303$   &  $0.424 $ \\ \hline
     \texttt{CatchOver2Underarm}  &  $0.366$        &  $0.166$   &  $0.466 $ \\ \hline
     \texttt{CatchAbreast}  &  $ 0.130$        &  $0.391$   &  $ 0.478 $ \\ \hline
     \texttt{HandOver}  &  $0.222$        &  $0.296$   &  $0.481 $ \\ \hline
     \texttt{BlockStack}  &  $0.280$        &  $0.280$   &  $ 0.440$ \\ \hline
     \texttt{CatchUnderarm}  &  $0.344$        &  $0.137$   &  $0.517 $ \\ \hline
     \texttt{BottleCap}  &  $0.212$        &  $0.212$   &  $ 0.575$ \\ \hline
     \texttt{LiftUnderarm}  &  $0.344$        &  $0.206$   &  $ 0.448$ \\ \hline
     \texttt{TwoCatchUnderarm}  &  $0.2$        &  $0.25$   &  $ 0.55$ \\ \hline
     \texttt{DoorOpenInward}  &  $0.176$        &  $0.205$   &  $0.617 $ \\ \hline
     \texttt{DoorOpenOutward}  &  $0.352$        &  $0.117$   &  $0.529 $ \\ \hline
     \texttt{DoorCloseInward}  &  $0.107$        &  $0.285$   &  $ 0.607$ \\ \hline
     \texttt{PushBlock}  &  $0.280$        &  $0.200$   &  $0.520 $ \\ \hline
     \texttt{Scissors}  &  $0.233$        &  $0.066$   &  $ 0.7$ \\ \hline
     \texttt{Pen}  &  $0.285$        &  $0.142$   &  $0.571 $ \\ \hline
     \texttt{GraspAndPlace}  &  $0.192$        &  $0.269$   &  $0.538 $ \\ \midrule
     Total  &  $0.251$        & $0.222$   & $0.527$ \\ \midrule
\end{tabular}
\end{table}

\begin{table}[htbp]
\caption{Preference results for each training task (final iteration).}
\label{tab:prefer_breakdown2}
\centering
\begin{tabular}{c|ccc}
\toprule
      Task &  \multirow{1}{*}{Policy+RM} & \multicolumn{1}{c}{Original Policy} & \multicolumn{1}{c}{Not Sure} \\ \hline
     \texttt{ShadowHand}  &  $0.310$        &  $0.068$   &  $0.620$ \\ \hline
     \texttt{Switch}  &  $0.321$        &  $0.142$   &  $0.535 $ \\ \hline
     \texttt{CatchOver2Underarm}  &  $0.263$        &  $0.263$   &  $0.473 $ \\ \hline
     \texttt{CatchAbreast}  &  $0.333$        &  $0.074$   &  $ 0.592 $ \\ \hline
     \texttt{HandOver}  &  $0.548$        &  $0.032$   &  $0.419$ \\ \hline
     \texttt{BlockStack}  &  $0.259$        &  $0.185$   &  $ 0.555$ \\ \hline
     \texttt{CatchUnderarm}  &  $0.178$        &  $0.107$   &  $0.714 $ \\ \hline
     \texttt{BottleCap}  &  $0.428$        &  $0.142$   &  $ 0.428$ \\ \hline
     \texttt{LiftUnderarm}  &  $0.320$        &  $0.080$   &  $ 0.6$ \\ \hline
     \texttt{TwoCatchUnderarm}  &  $0.269$        &  $0.230$   &  $ 0.5$ \\ \hline
     \texttt{DoorOpenInward}  &  $0.307$        &  $0.192$   &  $0.5 $ \\ \hline
     \texttt{DoorOpenOutward}  &  $0.481$        &  $0.148$   &  $0.370 $ \\ \hline
     \texttt{DoorCloseInward}  &  $0.363$        &  $0.090$   &  $0.545$ \\ \hline
     \texttt{PushBlock}  &  $0.481$        &  $0.074$   &  $0.444 $ \\ \hline
     \texttt{Scissors}  &  $ 0.250$        &  $0.107$   &  $ 0.642$ \\ \hline
     \texttt{Pen}  &  $0.740$        &  $0.0$   &  $0.259$ \\ \hline
     \texttt{GraspAndPlace}  &  $0.258$        &  $0.290$   &  $0.451 $ \\ \midrule
     Total  &  $0.357$        & $0.134$   & $0.509$ \\ \midrule
\end{tabular}
\end{table}

\begin{table}[htbp]
\caption{Preference results for each unseen task (with final iteration RM).}
\label{tab:unseen_prefer_breakdown}
\centering
\begin{tabular}{c|ccc}
\toprule
       Task &  \multirow{1}{*}{Policy+RM} & \multicolumn{1}{c}{Original Policy} & Not Sure  \\ \hline
     \texttt{CatchAbreastPen}  &  $0.327$        &  $0.086$  &  $0.586$\\ \hline
     \texttt{TwoCatchAbreast}  &  $0.250$        &  $0.148$ & $0.601$  \\ \hline
     \texttt{CatchUnderarmPen}  &  $0.219$        &  $0.134$ &  $0.645$ \\ \hline
     \texttt{GraspAndPlaceEgg}  &  $0.209$        &  $0.147$ & $0.643$ \\ \midrule
     Total  &  $0.249$        & $0.130$   &  $0.621$\\ \midrule
\end{tabular}
\end{table}

\section{Detailed Success Results}
\label{app:success}
The complete success rate results of each task as a detailed version of Tab.~\ref{tab:success} is shown in Tab.~\ref{tab:success_full}.

\begin{table}[htbp]
\caption{Breakdown success rates over seen tasks (17) and unseen tasks (4).}
\label{tab:success_full}
\begin{tabular}{c|cc}
\toprule
      \textbf{Seen Task} &  \multirow{1}{*}{\textbf{Policy+RM}} & \multicolumn{1}{c}{{Original Policy}}  \\ \hline
     \texttt{ShadowHand}  &  $0.40\pm{0.20} $        &  $0.98\pm0.94$   \\ \hline
     \texttt{Switch}  &  $0.00\pm0.00$        &  $0.00\pm0.00$   \\ \hline
     \texttt{CatchOver2Underarm}  &  $0.62\pm0.25$        &  $0.74\pm0.24$    \\ \hline
     \texttt{CatchAbreast}  &  $0.60\pm0.23$        &  $0.58\pm0.24$    \\ \hline
     \texttt{HandOver}  &  $0.79\pm 0.31$        &  $0.83\pm0.24$    \\ \hline
     \texttt{BlockStack}  &  $0.39 \pm 0.24$        &  $0.38\pm0.29$   \\ \hline
     \texttt{CatchUnderarm}  &  $0.56\pm0.28$        &  $0.68\pm0.23$   \\ \hline
     \texttt{BottleCap}  &  $0.58\pm0.39 $        &  $0.63\pm0.28$    \\ \hline
     \texttt{LiftUnderarm}  &  $0.00\pm0.00$        &  $0.10\pm0.14$    \\ \hline
     \texttt{TwoCatchUnderarm}  &  $0.05\pm0.05$        &  $0.04\pm0.07$   \\ \hline
     \texttt{DoorOpenInward}  &  $0.03\pm0.04$        &  $0.21\pm0.26$   \\ \hline
     \texttt{DoorOpenOutward}  &  $0.85\pm0.34$        &  $0.85\pm0.32$    \\ \hline
     \texttt{DoorCloseInward}  &  $0.5\pm0.46$        &  $0.54\pm0.47$   \\ \hline
     \texttt{PushBlock}  &  $0.46\pm0.37$        &  $0.51\pm0.40$    \\ \hline
     \texttt{Scissors}  &  $0.85\pm0.34$        &  $0.9\pm0.27$   \\ \hline
     \texttt{Pen}  &  $0.78\pm0.29$        &  $0.66\pm0.43$   \\ \hline
     \texttt{GraspAndPlace}  &  $0.61\pm0.36$        &  $0.82\pm0.23$   \\ \hline
     \textbf{Seen Total}  &  $0.47$        & $0.55$   \\ \hline \hline
     \textbf{Unseen Task} &  \multirow{1}{*}{\textbf{Policy+RM}} & \multicolumn{1}{c}{{Original Policy}}  \\ \hline
     \texttt{CatchAbreastPen}  &  $0.52\pm{0.37} $        &  $0.51\pm0.35$   \\ \hline
     \texttt{CatchUnderarmPen}  &  $0.70\pm0.36$        &  $0.67\pm0.37$    \\ \hline
     \texttt{GraspAndPlaceEgg}  &  $0.65\pm0.35$        &  $0.74\pm0.29$   \\  \hline
     \texttt{TwoCatchAbreast}  &  $0.00\pm0.00$        &  $0.00\pm0.00$   \\ \hline
     \textbf{Unseen Total}  &  $0.47$        & $0.48$   \\ \bottomrule
\end{tabular}
\end{table}

\newpage
\section{Real Robot Experiments}
\label{app:real_exp}

\begin{figure}[htbp]
\centering\includegraphics[width=\textwidth]{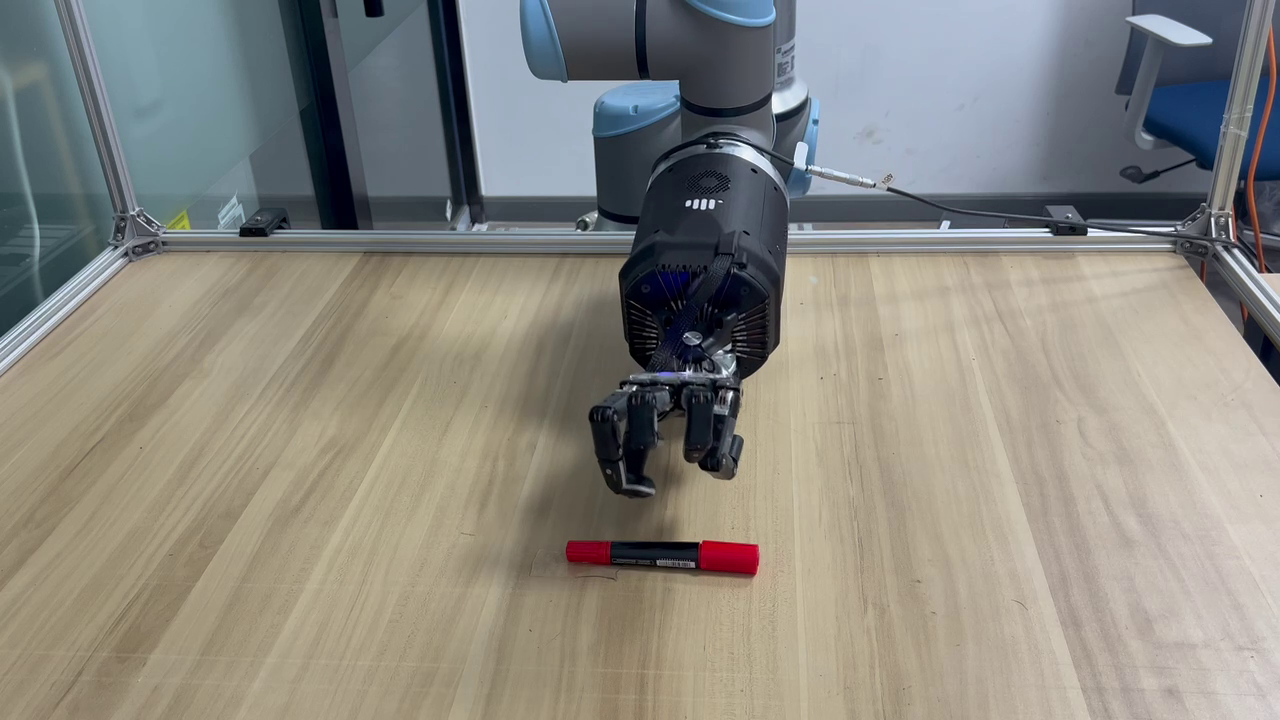}
	\caption{Real-robot experiment setup includes a Shadow Hand, a UR10e robotic arm and the objects to interact with. A camera is located in front to capture the trajectories of the robot movement.}
	\label{fig:real_setup}
\vskip -0.1in
\end{figure}
\begin{figure*}[htbp]
    \includegraphics[width=\columnwidth]{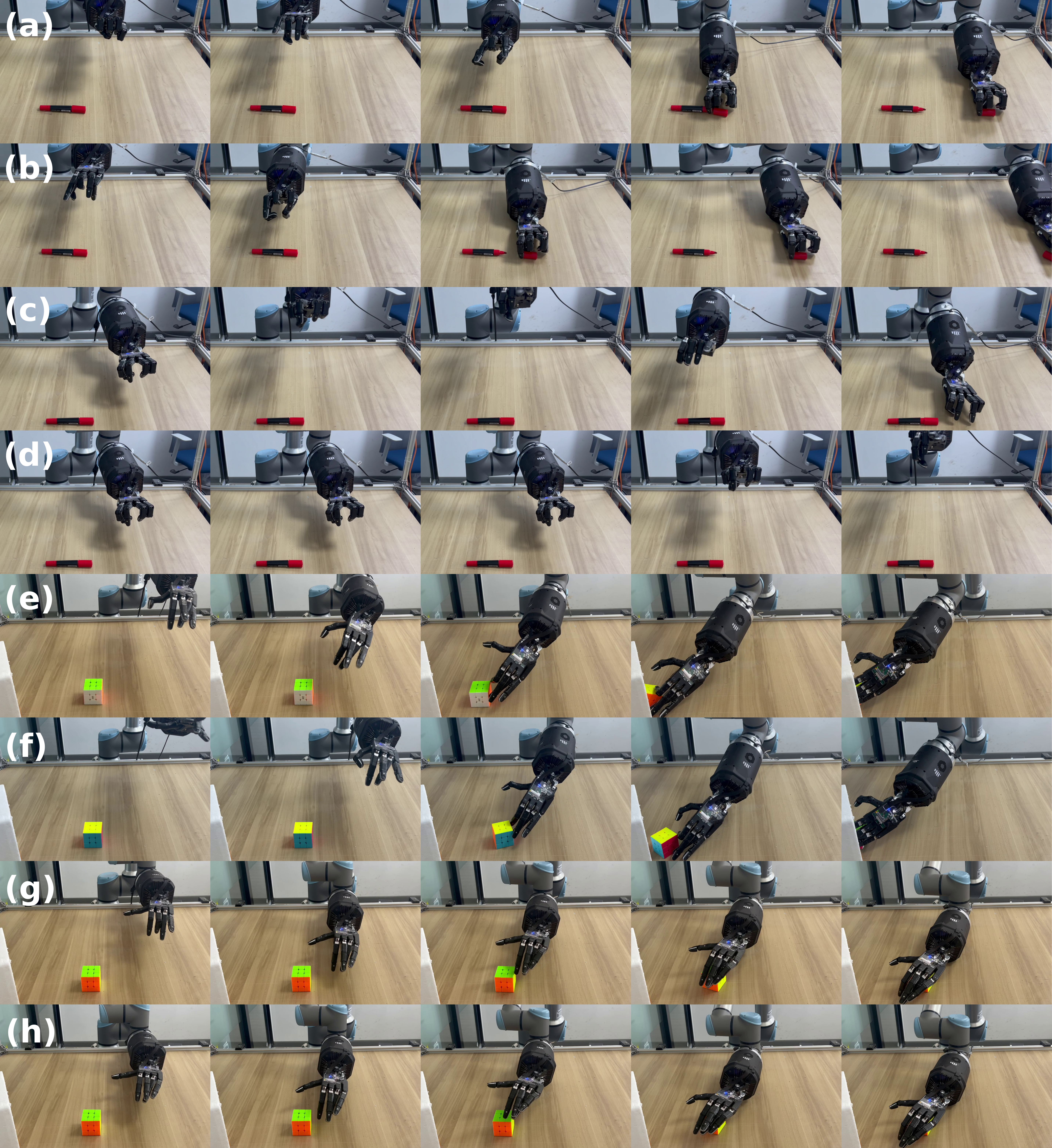}
	\caption{The figure displays the trajectories of simulation-to-reality experiments with RM fine-tuning (ours) and vanilla policy training (baseline) on two tasks \texttt{Pen} and \texttt{Relocate}. [a-d] are \texttt{Pen} and [e-h] are \texttt{Relocate}. [a-b] and [e-f] are results by our method while [c-d] and [g-h] are the baseline results. From left to right it shows the procedure of the task completion.}
	\label{fig:detail_real_exp}
\end{figure*}

The setup of real-robot experiments includes a Shadow Hand mounted at the end of the UR10e robotic arm, as shown in Fig.~\ref{fig:real_setup}. The reason for using a single Shadow Hand and one robotic arm instead of two is the hardware limitation. However, this will not affect our demonstration purpose, given the fact that by design the RM trained in our method can work for two hands separately, which also allows it tune polices for a single hand directly. Both the Shadow Hand and the robotic arm are controlled simultaneously at a frequency of 10Hz. Two tasks \texttt{Pen} and \texttt{Relocate} are evaluated in the experiments. Fig.~\ref{fig:detail_real_exp} displays the trajectories of robots following policies trained with and without RM fine-tuning process. We show two trials for each setting. From the results we can see that, the fine-tuned policies demonstrate more human-like behavior especially on the task \texttt{Pen}. Fine-tuned polices also have better task completion performances in these two real-world tasks. For \texttt{Pen}, the hand is able to unplug the cap of the pen (with the pen itself anchored on the table) and move it away. For \texttt{Relocate}, the hand is able to push the cube away from its original position towards the target position in the white box located at the bottom left corner of the frame.
However, we would like to additionally declare that the task completion performance is not guaranteed to be improved by the fine-tuning approach for general tasks. These experiments show that in some cases fine-tuning can help. It remains to be investigated about how to improve the task completion together with generating more human-like behavior.

\newpage
\section{Lessons for Reward Model}
\label{sec:caveats}
During the training of RM and policy fine-tuning, we find that some detailed choices can significantly affect the final performances. Therefore, we emphasize the subtleties in using RM for tuning behaviors of robot policies. This section also serves as ablation study for our design choices.

\subsection{Objectives}
In Sec.~\ref{sec:policy_rm}, we introduce the objective used for tuning the robot task-specific policies with the RM (Eq.~\eqref{eq:rm_obj1}), and we compare it with an alternative objective (second) for RLHF commonly used in other works~\cite{jaques2019way, stiennon2020learning} or its variant~\cite{ouyang2022training} which is commonly used in fine-tuning LLM with RLHF: 
\begin{equation}
    \mathcal{J}(\pi_\theta)=\mathbb{E}_{a_t\sim\pi_\theta(\cdot|s_t)}\big[\sum_{t=1}^T \big(r_\text{HF}(\tau_t)-\beta \log \frac{\pi_\theta(a_t|s_t)}{\pi_{\theta_0}(a_t|s_t)}\big)\big]
\label{eq:rm_obj2}
\end{equation}
This alternative objective does not work well in our settings. In previous work~\cite{ouyang2022training, jaques2019way,stiennon2020learning} using Eq.~\eqref{eq:rm_obj2} as objective, the role of RM is to further improve the policy performance on the natural language tasks, where the RM aligns well with the policy training objective. Here in our human-like behavior setting, the objective of RM and the objective of original task completion has a certain level of divergence. Directly tuning polices with Eq.~\eqref{eq:rm_obj2} will lead to a high accumulative $r_\text{HF}$ but poor performance of task completion, which is not desirable.

\begin{figure}[h]
	\centering\includegraphics[width=0.48\columnwidth]{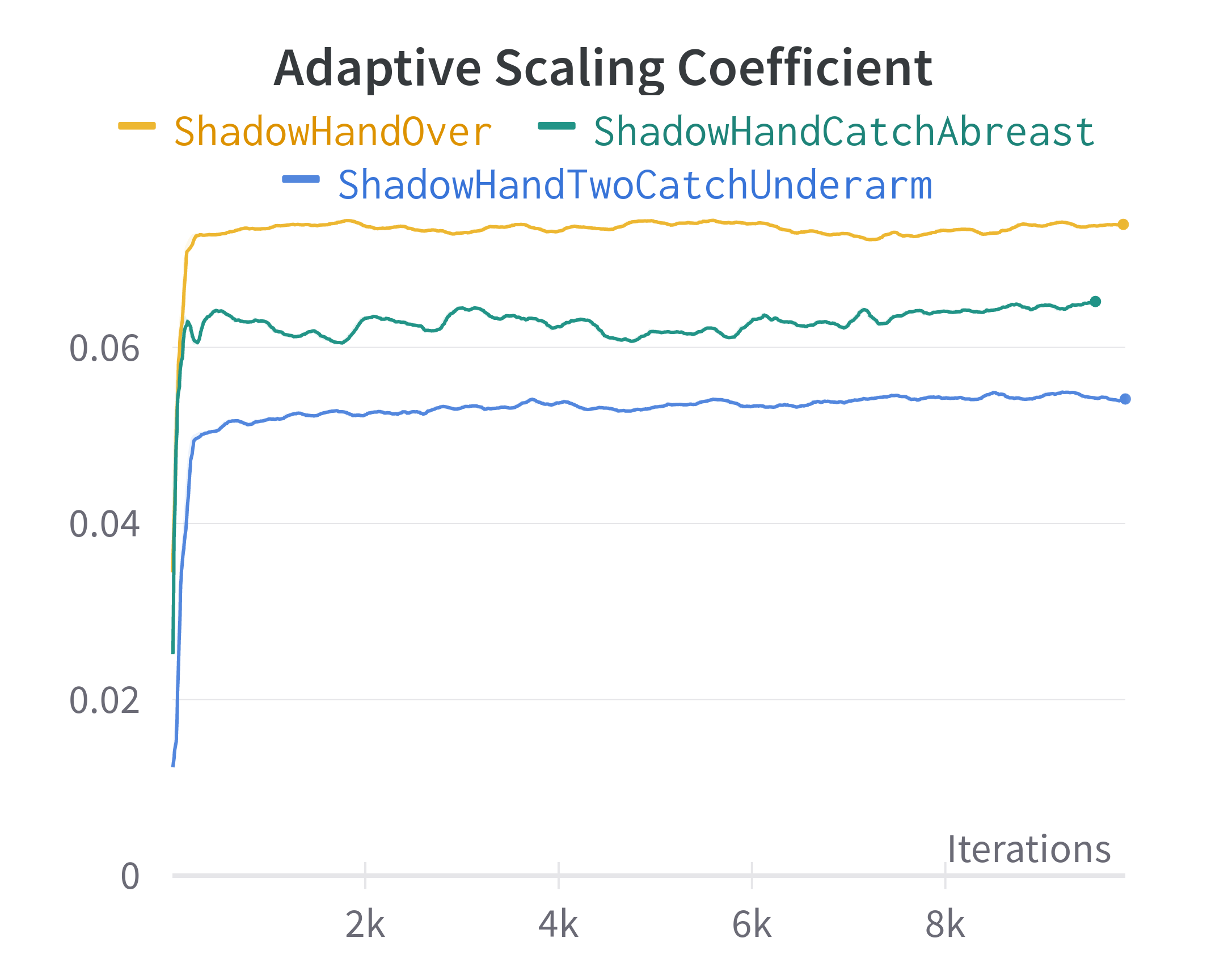}
 \centering\includegraphics[width=0.48\columnwidth]{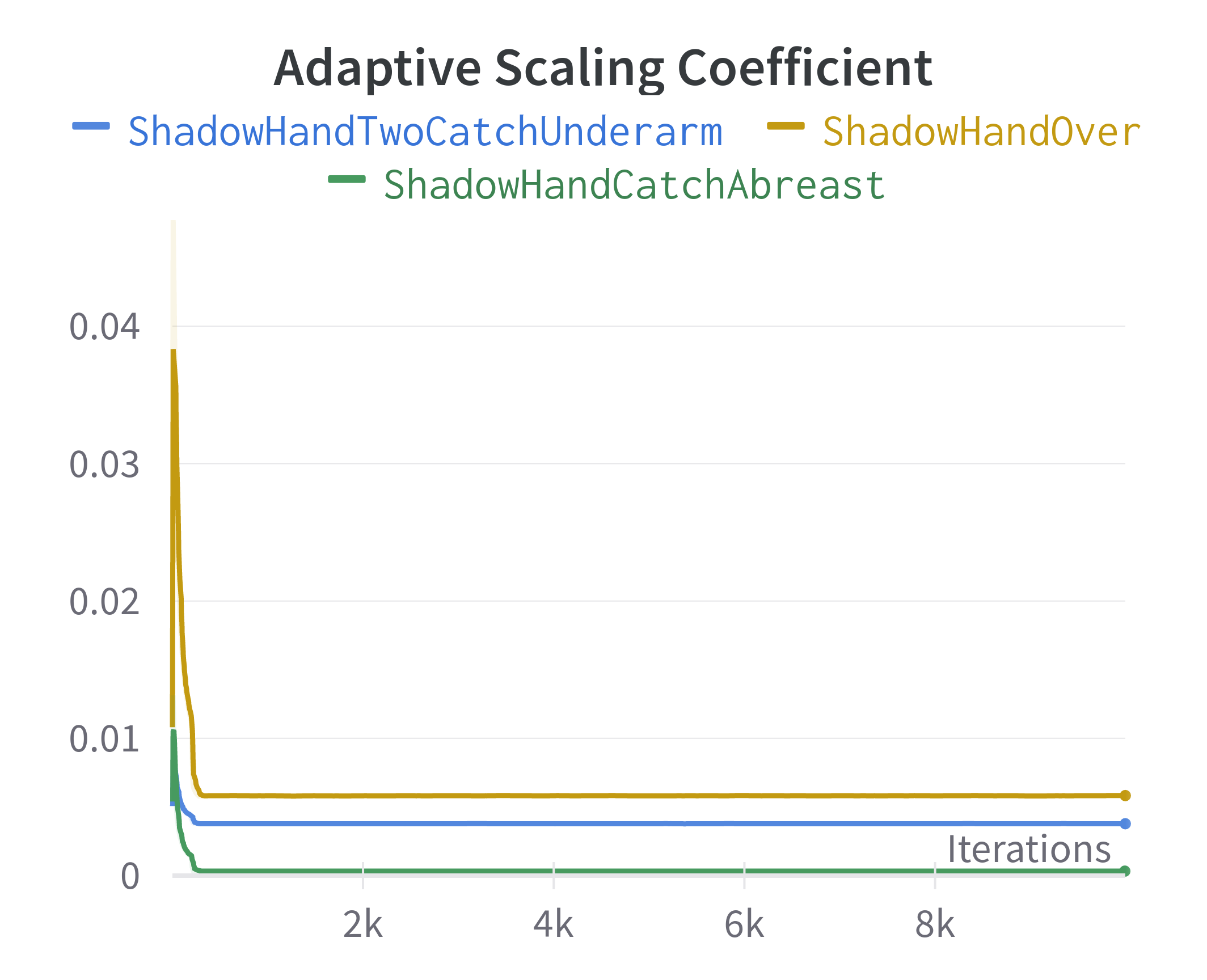}\\
	\centering\includegraphics[width=0.48\columnwidth]{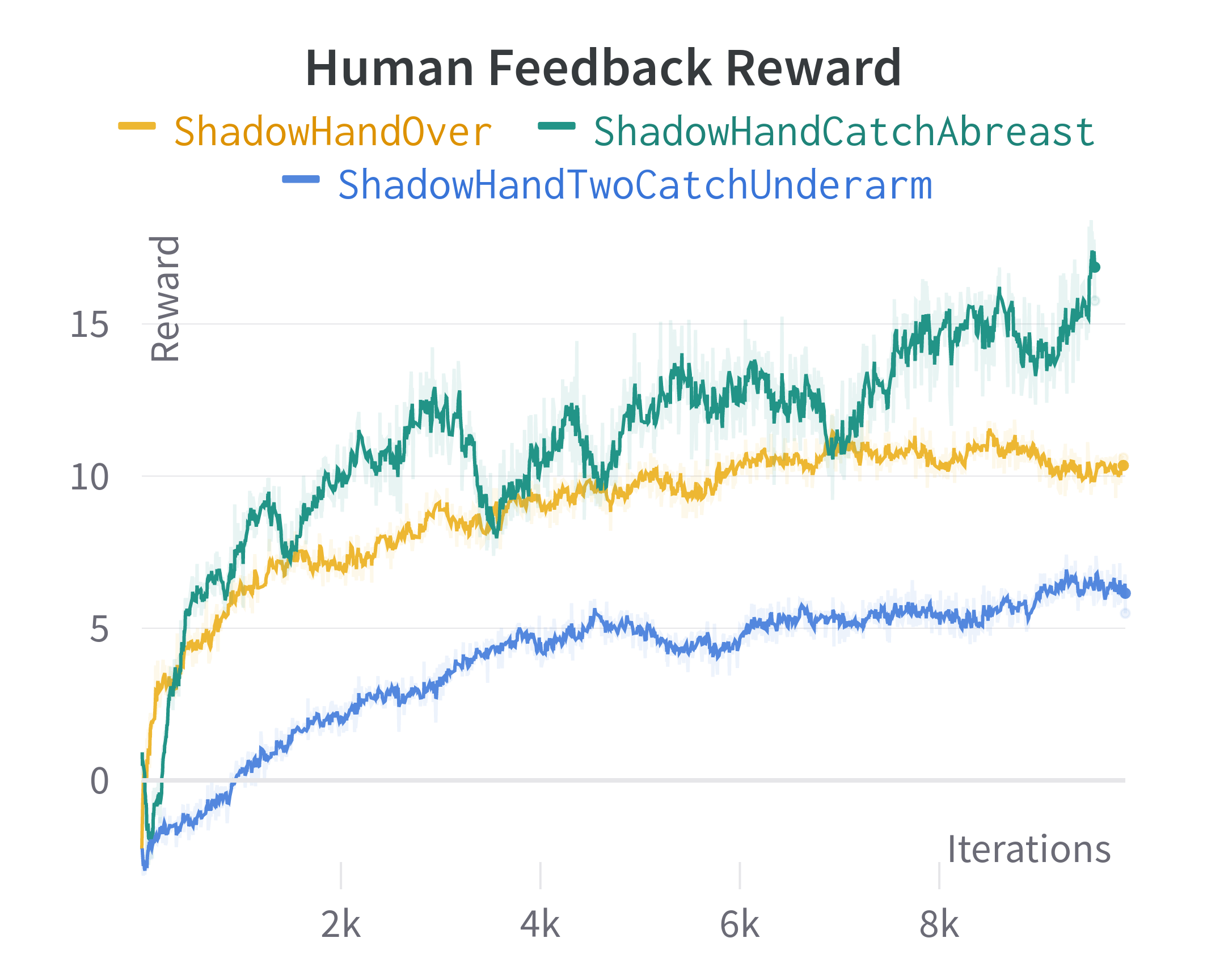}
 \centering\includegraphics[width=0.48\columnwidth]{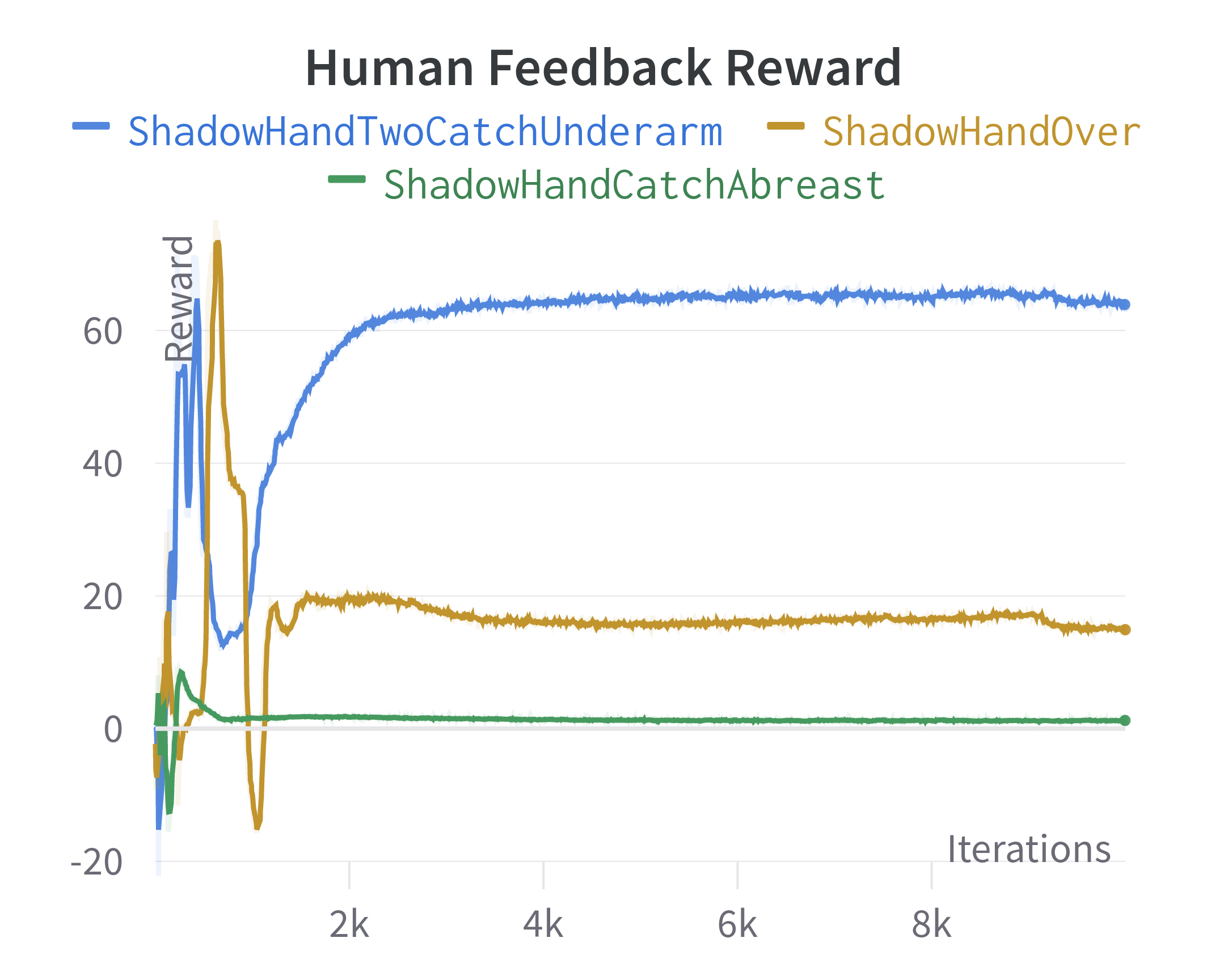}\\
	\centering\includegraphics[width=0.48\columnwidth]{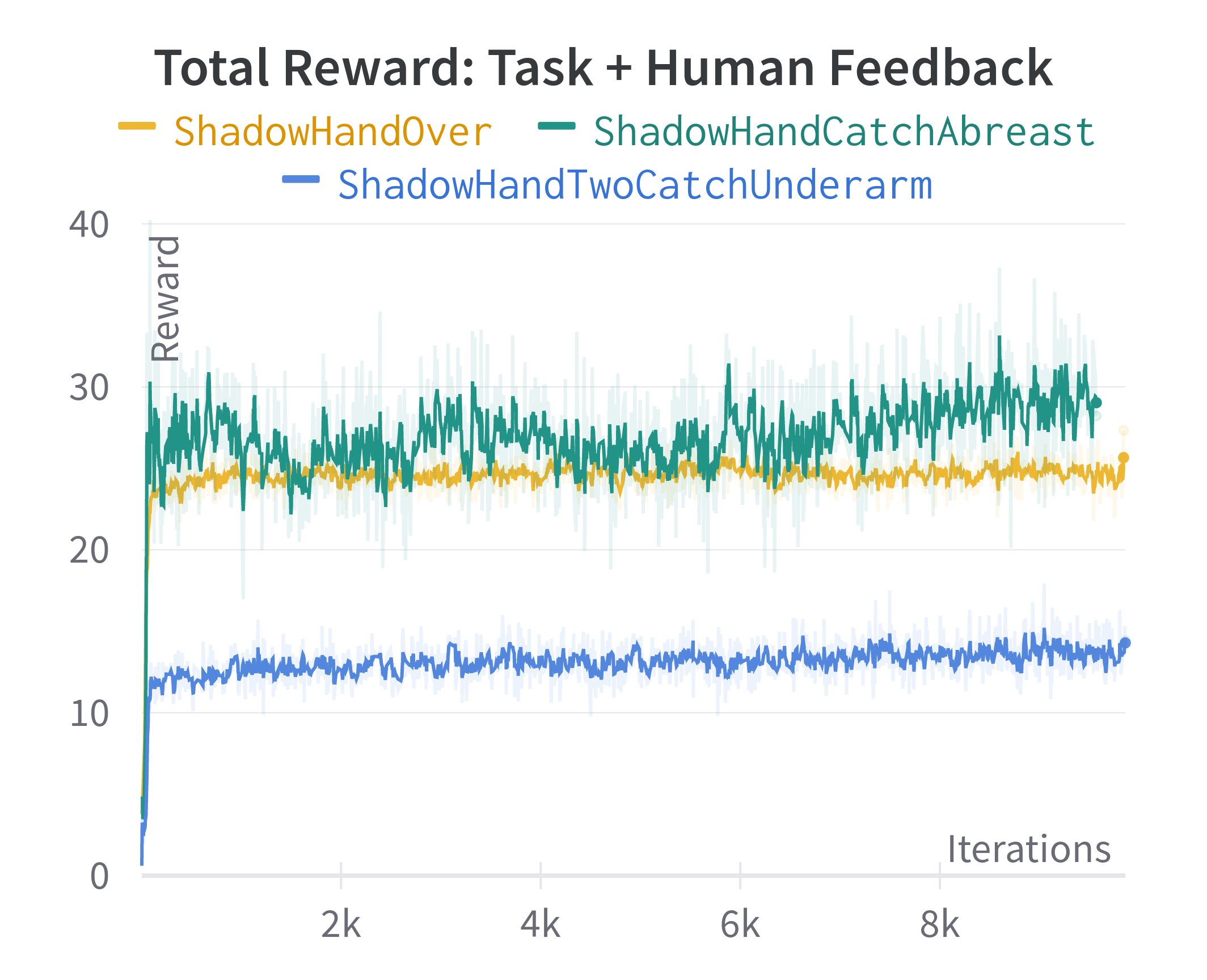} 
	\centering\includegraphics[width=0.48\columnwidth]{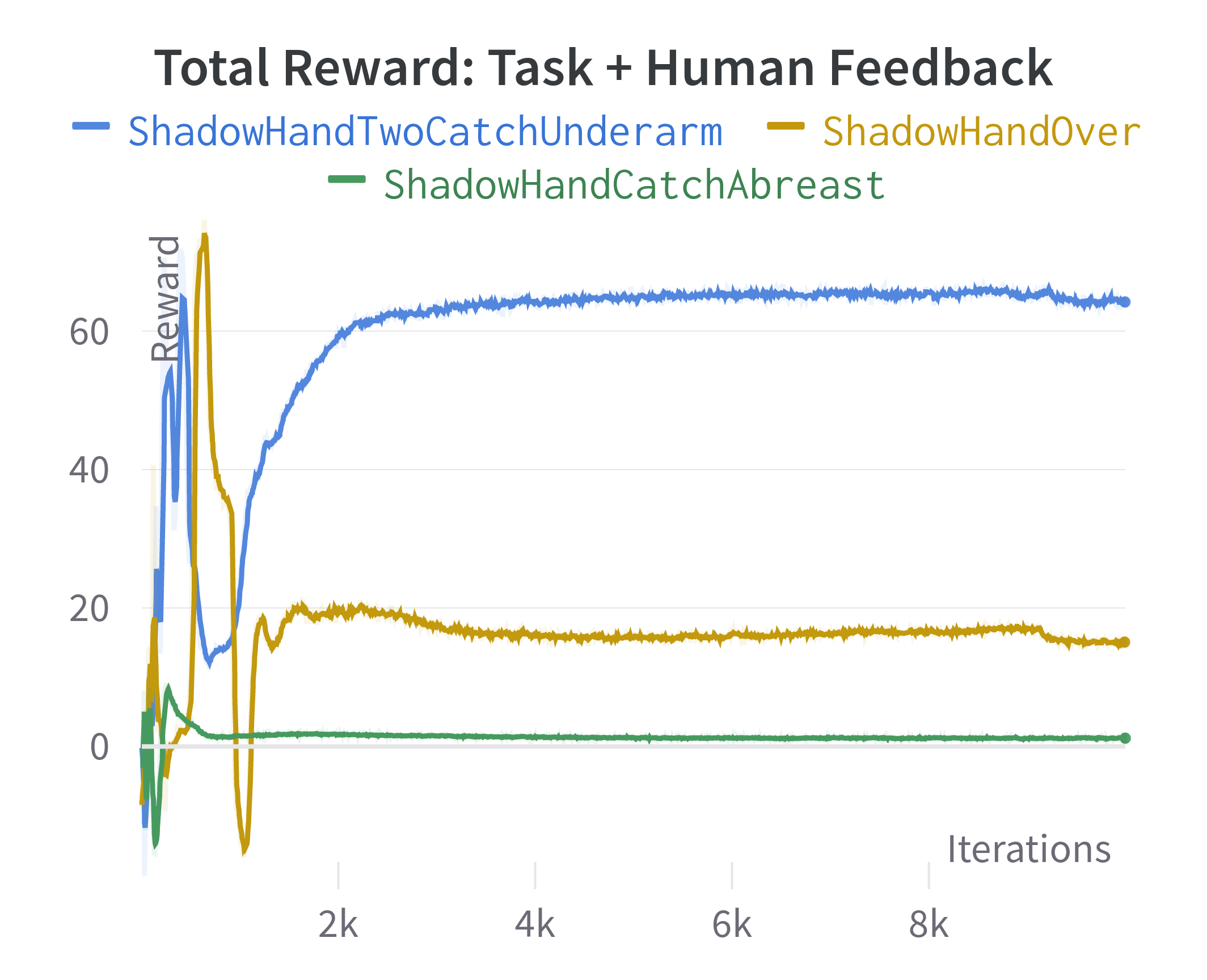}
	\caption{The learning curves of policy fine-tuning with the RM on three tasks. Three diagrams on the left uses the first objective as Eq.~\eqref{eq:rm_obj1}; Three diagrams on the right uses the second objective as Eq.~\eqref{eq:rm_obj2}. }
	\label{fig:rm_policy}
\end{figure}

In Fig.~\ref{fig:rm_policy} we compare the two objectives for three different tasks: \texttt{HandOver}, \texttt{CatchAbreast} and \texttt{TwoCatchUnderarm}. The adaptive scaling coefficient is $c$ in Eq.~\eqref{eq:rm_obj1}, which is also the smoothed value of the task reward. For the second objective, although there is no such coefficient in the equation, we also track this value as an indicator of the task completion during training. We take $\alpha=0.2$ and $\beta=20.0$ for this experiment\footnote{We also test with larger $\beta$ values and the results are similar.}. The results show that although the second objective optimizes the policy with higher human feedback rewards, it severely hurts the task completion performance in the process of policy tuning, since the task rewards are very small and not increasing during the training. Under the first objective, the policies manage to improve the reward for both the task completion and human preference. 

\subsection{Scaling matters} 
\begin{figure}[htbp]
	\centering\includegraphics[width=0.48\columnwidth]{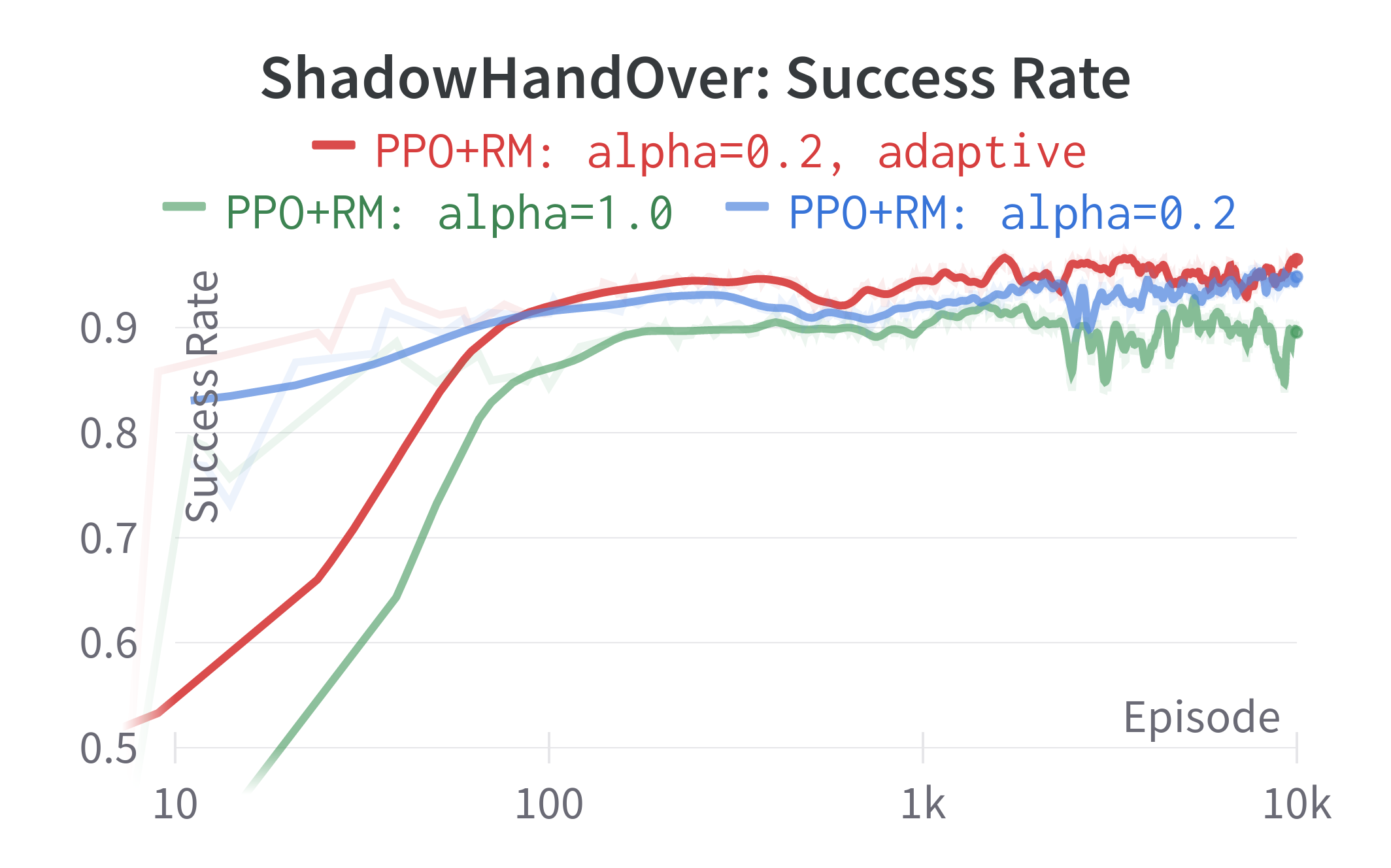}
	\centering\includegraphics[width=0.48\columnwidth]{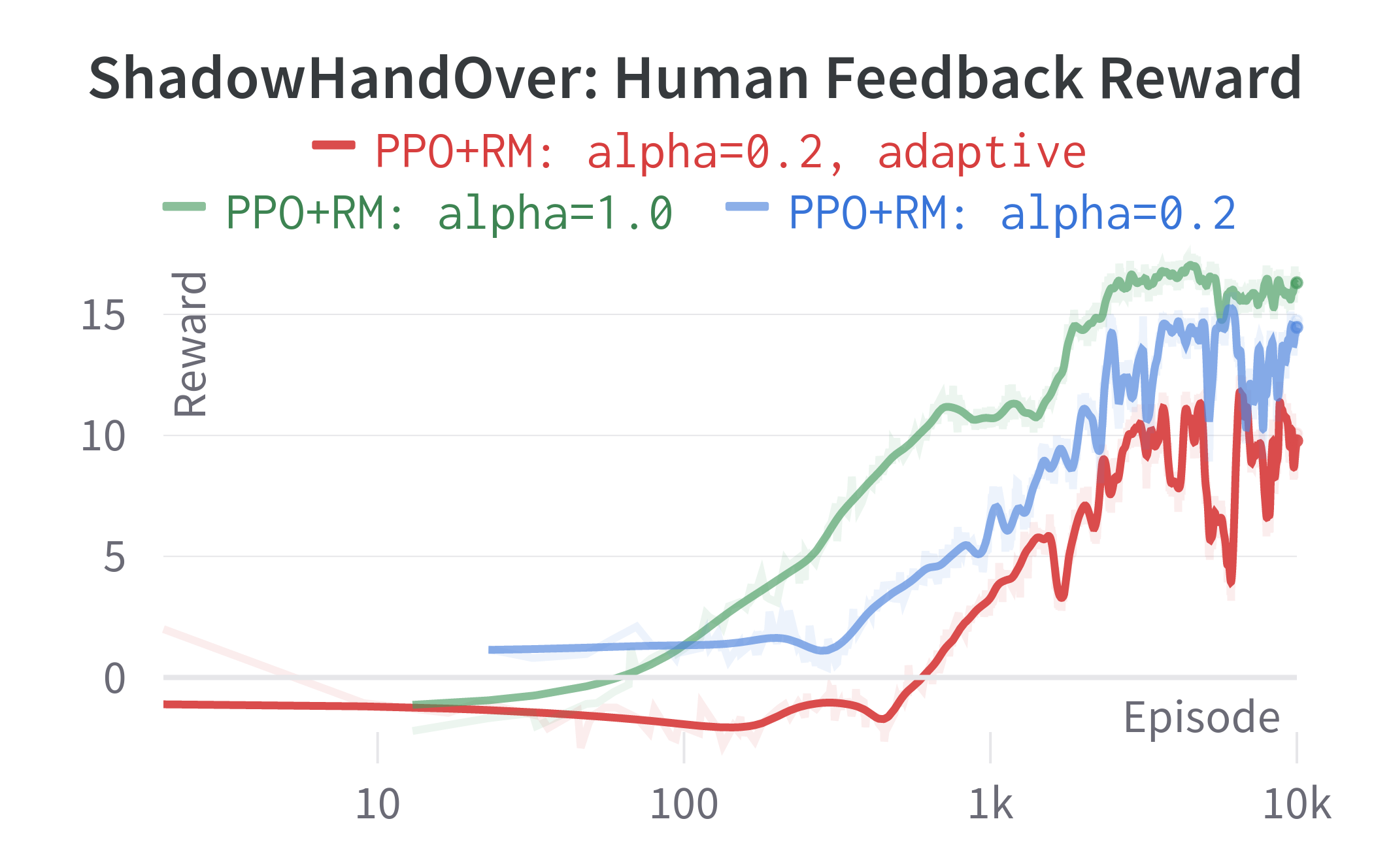}\\
	\centering\includegraphics[width=0.48\columnwidth]{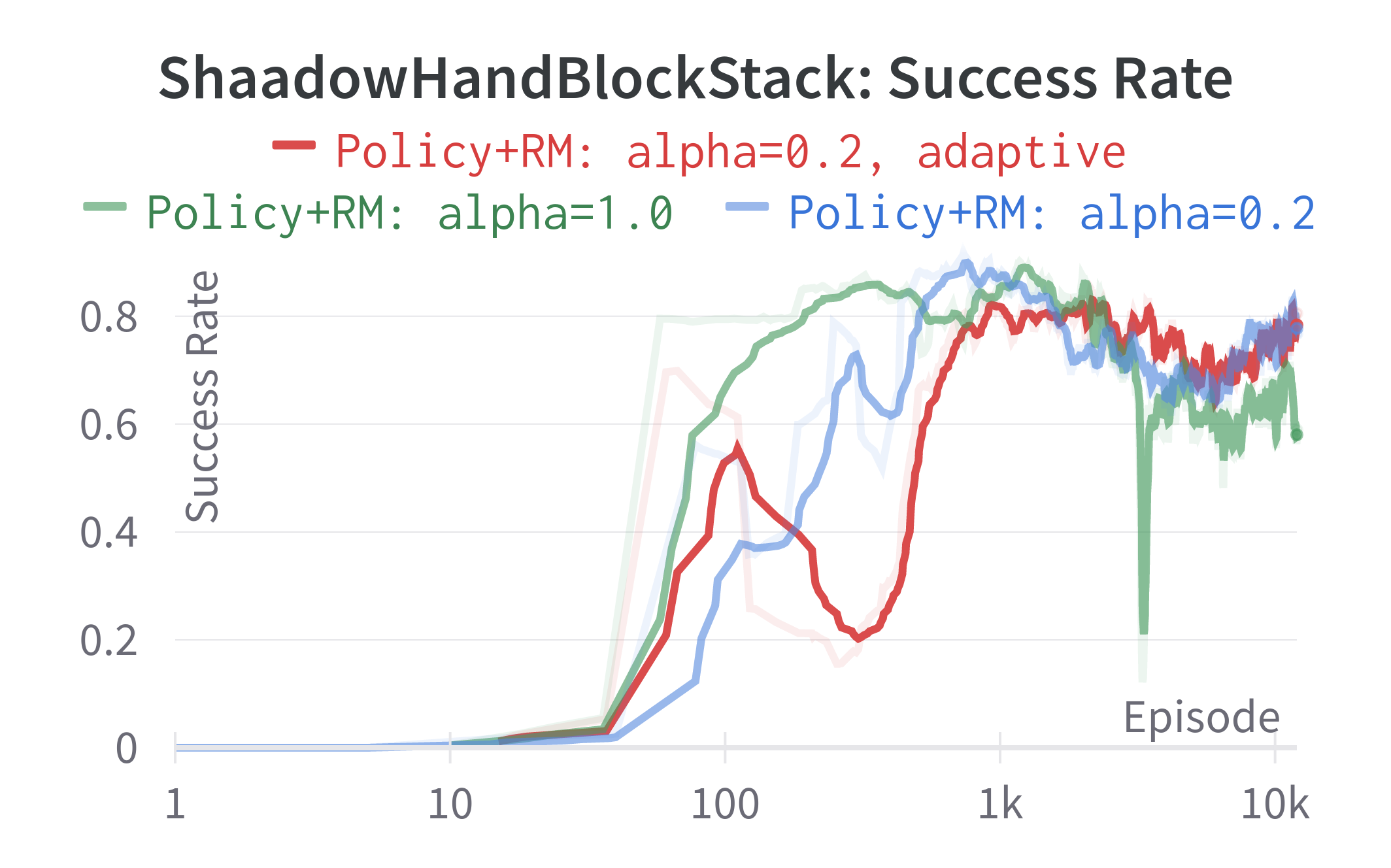}
	\centering\includegraphics[width=0.48\columnwidth]{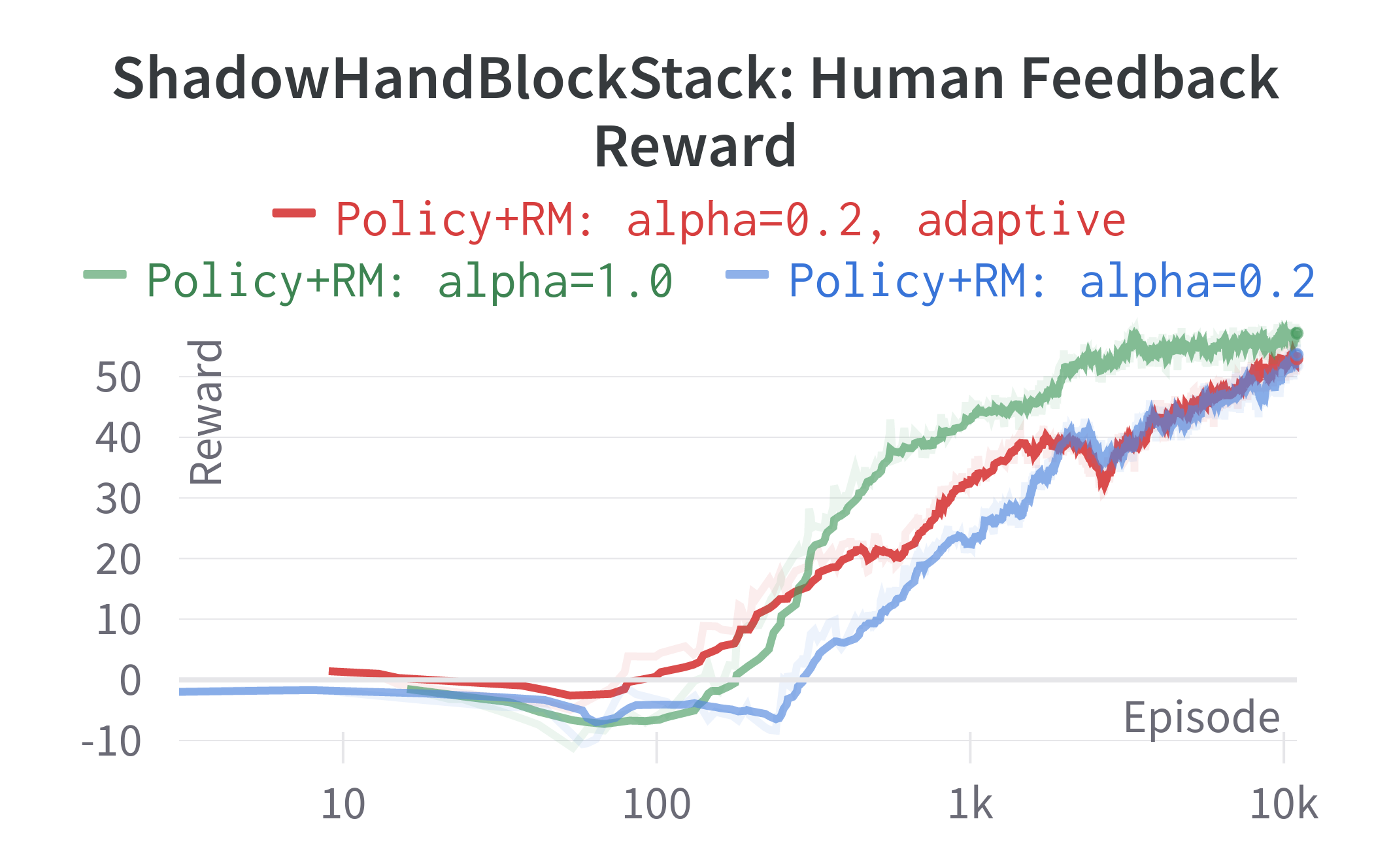}\\
	\centering\includegraphics[width=0.48\columnwidth]{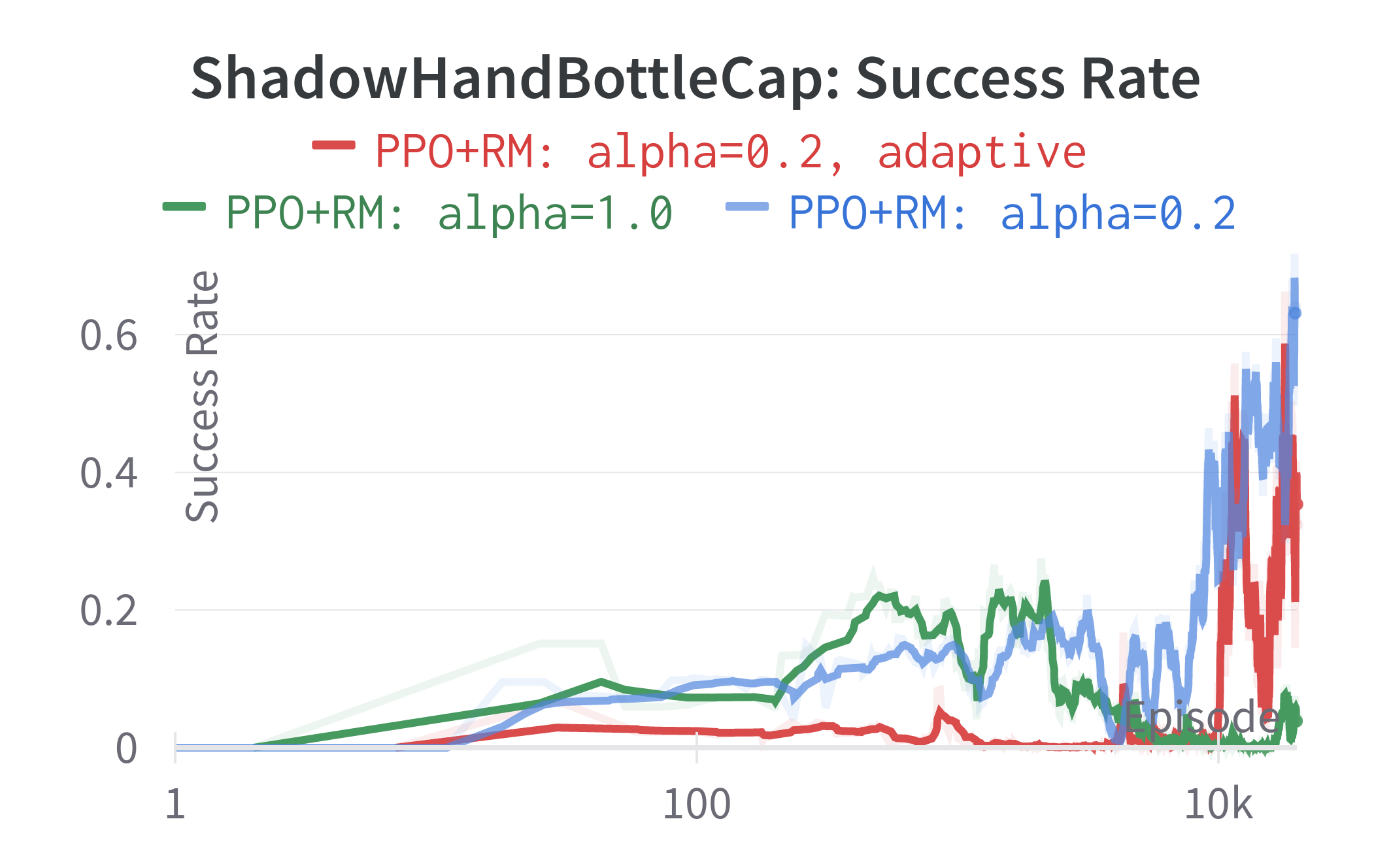}
	\centering\includegraphics[width=0.48\columnwidth]{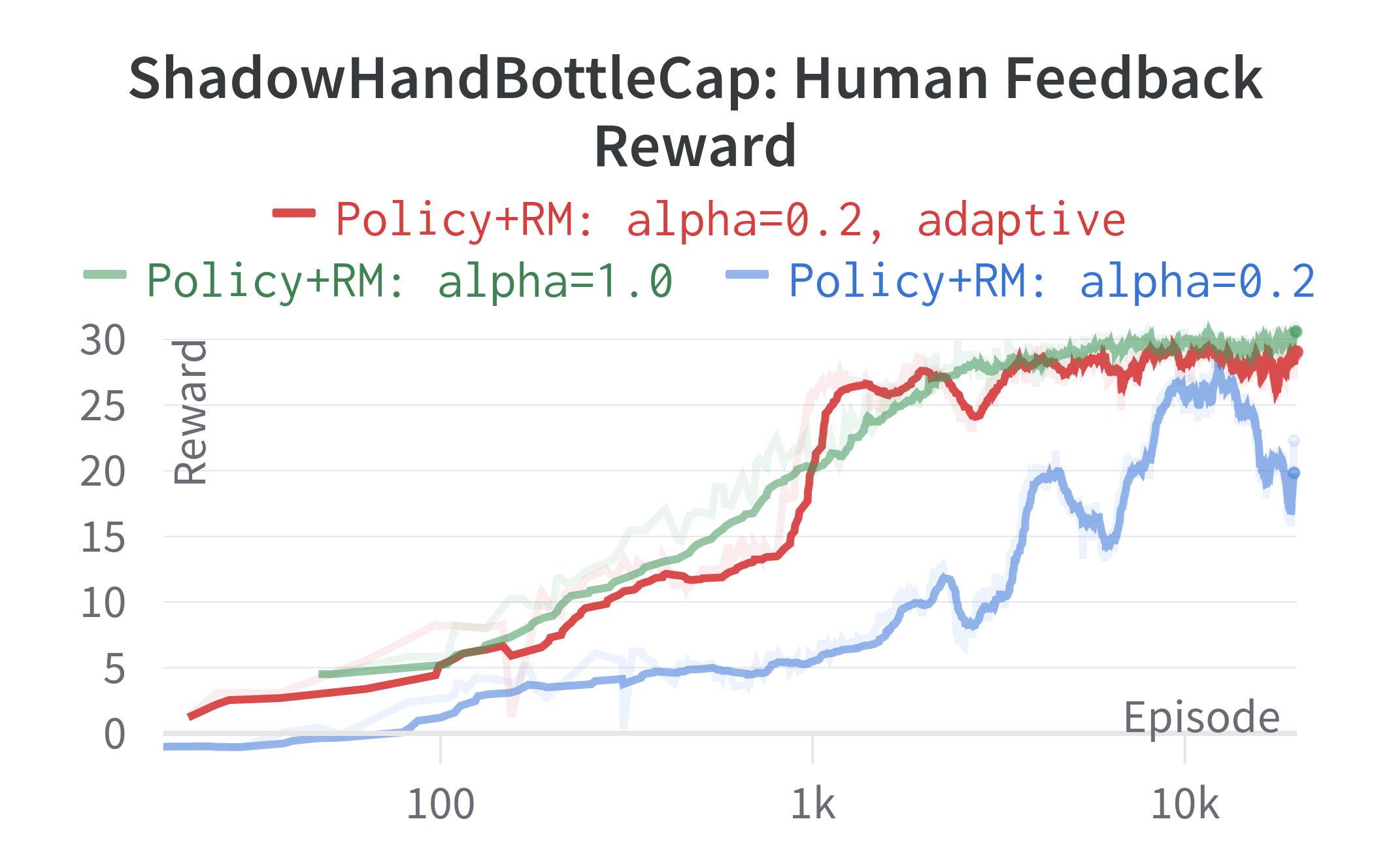}
	\caption{The learning curves of policy fine-tuning with the RM on three tasks. Top: \texttt{HandOver}; Bottom: \texttt{CatchAbreast}. }
	\label{fig:rm_scale}
\end{figure}

Following the objective in Eq.~\eqref{eq:rm_obj1} for policy tuning, we further notice that the scaling coefficients $\alpha$ and $c$ can affect the policy performance significantly. 
Without the proper scaling of the human preference reward, the performance of task completion can also be bad. In Fig.~\ref{fig:rm_scale}, we compare the task success rates and episodic rewards for different scaling choices on three tasks: \texttt{HandOver}, \texttt{BlockStack}, \texttt{BottleCap}. Specifically, whether using the adaptive coefficient $c$ ($c=1$ if not adaptive) is chosen and the value of $\alpha$ are the hyperparameters in this experiment. Although the effects are not as significant as the change of optimization objective in above section, it also shows that using the adaptive coefficient with a smaller scaling factor $\alpha$ usually leads to higher task success rates as well as relatively high human feedback rewards, which indicates the better policy performance in terms of both task completion and human-like behaviors.


\subsection{The RM may not serve to guide the task completion}
For some hard tasks like \texttt{Kettle} and \texttt{DoorCloseOutward}, the PPO algorithm with certain entropy bonus for boosting exploration is not sufficient to acquire the optimal policy in terms of task completion. In this case, RM with human-like behavior as the criterion is unlikely to help with the task completion even after many iterations of policy fine-tuning. 
 
\subsection{The RM can be hard to achieve human-like behavior in special cases} For some tasks like \texttt{SwingCup}, it could be hard for PPO policies to explore human-like behaviors even with the proper specification of initial pre-grasp poses and sophisticated reward engineering. In this case, the diverse polices to collect human feedback do not even contain any human-like behavior. It is can be very hard to use the RM approach to get the human-like behavior after many iterations of fine-tuning.

\end{document}